\documentclass[11pt, table]{article}

\usepackage[preprint]{acl}

\usepackage{times}
\usepackage{latexsym}

\usepackage[T1]{fontenc}

\usepackage[utf8]{inputenc}

\usepackage{microtype}

\usepackage{inconsolata}

\usepackage{graphicx}

\usepackage{nicematrix, booktabs}
\usepackage{amsmath}
\usepackage{amsfonts}
\usepackage{xcolor}
\usepackage{multirow}
\usepackage{listings}
\usepackage{stfloats} 
\usepackage{tcolorbox}
\usepackage{makecell}
\usepackage{enumitem}
\usepackage{upquote} 
\usepackage{arydshln} 
\usepackage{circledsteps}
\usepackage{soul}
\usepackage{tikz}
\usepackage[autostyle]{csquotes}

\newcommand*\myCircled[1]{\tikz[baseline=(char.base)]{
            \node[shape=circle,fill,inner sep=1pt] (char) {\textcolor{white}{#1}};}}

\newcommand{\p}{\mathcal{P}}

\newcommand{\C}{\mathcal{C}}
\newcommand{\Cc}{\mathcal{C}'}
\newcommand{\M}{\mathcal{M}}

\newcommand{\X}{\mathbf{X}}

\newcommand{\llamaModel}{Llama3.1 }

\lstdefinestyle{mystyle}{
    language=Python,
    basicstyle=\ttfamily\footnotesize,
    keywordstyle=\color{blue},
    commentstyle=\color{gray},
    stringstyle=\color{red},
    breaklines=true,
    breakatwhitespace=true,
    frame=single
}

\title{Systematic Evaluation of Optimization Techniques for \\
Long-Context Language Models}

\iftrue
\author{
 \textbf{Ammar Ahmed\textsuperscript{*, 1}}
 \textbf{Sheng Di\textsuperscript{2}},
 \textbf{Franck Cappello\textsuperscript{2}},
 \textbf{Zirui Liu\textsuperscript{1}},
 \textbf{Jingoo Han\textsuperscript{3}},
 \textbf{Ali Anwar\textsuperscript{1}}
\\
\\
 \textsuperscript{1}University of Minnesota, Twin Cities, USA \\
 \textsuperscript{2}Argonne National Labratory, Lemont, USA
 \textsuperscript{3}Samsung Semiconductor Inc.,
\\
\{ahme0599, zrliu, aanwar\}@umn.edu, 
sdi1@anl.gov, cappello@mcs.anl.gov, \\
jingoo.han@samsung.com
}
\fi

\begin{document}
\maketitle
\begingroup\def\thefootnote{*}\footnotetext{University of Minnesota. Correspondence to: Ammar Ahmed <ahme0599@umn.edu>}\endgroup
\begin{abstract}
Large language models (LLMs) excel across diverse natural language processing tasks but face resource demands and limited context windows. Although techniques like pruning, quantization, and token dropping can mitigate these issues, their efficacy in long-context scenarios and system evaluation remains underexplored. This paper systematically benchmarks these optimizations, characterizing memory usage, latency, and throughput, and studies how these methods impact the quality of text generation. We first analyze individual optimization methods for two LLM architectures supporting long context and then systematically evaluate combinations of these techniques to assess how this deeper analysis impacts performance metrics. We subsequently study the scalability of individual optimization methods on a larger variant with 70 billion-parameter model. Our novel insights reveal that naive combination inference optimization algorithms can adversely affect larger models due to compounded approximation errors, as compared to their smaller counterparts. Experiments show that relying solely on F1 obscures these effects by hiding precision–recall trade‑offs in question answering tasks. By integrating system-level profiling with task-specific insights, this study helps LLM practitioners and researchers explore and balance efficiency, accuracy, and scalability across tasks and hardware configurations.
\end{abstract}


\section{Introduction}

Large Language Models (LLMs) have become transformative tools in scientific computing, driving advances in areas such as computational physics, materials discovery, and automated code optimization~\cite{chaturvedi2024hpc}. Recent initiatives, including Argonne's AuroraGPT project~\cite{cappelloauroragpt}, aim to train trillion-parameter foundation models on DOE's exascale supercomputers. Empirical evidence shows that larger LLMs deliver stronger generalization and task performance than smaller ones~\cite{zhang2022opt, touvron2023llama2, llama3modelcard}, partly due to the memorization capacity of transformer architectures, where increased attention heads and dimensionality offer richer contextual representations~\cite{mahdavi2023memorization}. 

However, scaling LLMs is resource-intensive, with significant memory requirements that restrict deployment flexibility and scope. One of the most important use cases for LLMs is processing and retaining large amounts of information that is time-consuming for humans. \textit{As HPC efforts push toward trillion-parameter models, effectively managing long context windows becomes a pivotal challenge for harnessing LLMs in real-world SC scenarios.} A key bottleneck in LLMs is their limited context windows, which confine processing to limited-length token sequences (e.g., 2048 for OPT~\cite{zhang2022opt}, 8192 for Llama-3~\cite{llama3modelcard}) and impede the leveraging of long-range dependencies. Although emerging models like \llamaModel\cite{grattafiori2024llama} and Mistral-Nemo support context windows up to 128K tokens, deploying such capabilities remains difficult due to prohibitive memory and compute overheads.

While optimization techniques such as quantization~\cite{lin2024awq, frantar2022gptq, xiao2023smoothquant} and pruning~\cite{sreenivas2024llm} reduce LLM resource demands, their adoption in long-context applications remains fraught with unresolved challenges. Existing studies~\cite{yuan-etal-2024-kv, jin-etal-2024-comprehensive, NEURIPS2024_028fcbcf} predominantly evaluate these optimization methods in isolation on short-context tasks, ignoring the complex interplay of memory, compute, and attention patterns inherent to long sequences (e.g., 32K–128K tokens).They  also largely omit analysis of how such methods interact, particularly in long-context inference workloads. For instance, our analysis shows that pruning techniques like those implemented in Minitron~\cite{minitron2024} preserve accuracy on question-answering tasks but significantly degrade performance on summarization tasks.

Compounding this issue is that hardware efficiency often leads to lack of practical validity.
While simulated optimizations promise theoretical speedups~\cite{NEURIPS2023_0df38cd1}, real-world gains depend on specialized kernels for sparse or low-precision operations—components that are rarely supported in production frameworks~\cite{frantar2022gptq}. Even when such kernels exist, variable factors like batch size and input length yield inconsistent performance, rendering reported improvements unreliable for deployment planning~\cite{frantar2024marlin}. For example, 4-bit quantization might reduce latency by 40\% for 8K-token inputs but deliver diminishing returns at 64K tokens due to memory bandwidth saturation. 

Optimizations that work well individually may not combine effectively. 
For example, quantizing a pruned model does not automatically yield extra performance. Quantization alone gives the lowest memory footprint and highest tokens-per-second throughput, but stacking it on top of pruning can actually reduce throughput. This suggests that inference optimizations are not necessarily additive and may conflict when combined naively, resulting in cascading approximation errors that compound exponentially across thousands of tokens. These disparities underscore a critical oversight: optimizations are rarely benchmarked in combination under realistic long-context workloads, leaving their interaction effects largely unexplored until deployment. Moreover, techniques successful for smaller models (e.g., 8B parameters) may not scale to larger models (70B or more). Quantization, for instance, can be highly effective on small models but may degrade performance with increasing model size. This paper studies these key points through a systematic, in-depth analysis of optimization strategies for LLMs in long-context scenarios. Our key contributions are:
\vspace{-2pt}
\begin{itemize}[leftmargin=2mm]

\item \textbf{System-Level Performance Characterization:} We profile memory consumption, latency, and throughput across context lengths (up to 45k tokens), quantifying how optimizations affect hardware utilization. This includes novel insights into the relationship between quantization levels, sparsity patterns, and memory bandwidth constraints.

\item \textbf{Comprehensive Method Comparison:} By testing hybrid approaches (e.g., quantized + pruned models), we reveal both synergistic and antagonistic effects. Our findings show that while 4-bit quantization and pruning techniques like Minitron perform well independently, their naive combination can lead to severe degradation in quality of text generated, achieving the worst average performance score across tasks.

\item \textbf{Evaluation Metric Insights:} We dissect QA metrics, demonstrating that high aggregate F1 scores often mask precision-recall imbalances. Our analysis reveals that different optimization techniques can disproportionately affect different aspects of model performance, particularly in tasks with complex dependencies like summarization versus question-answering.

\item \textbf{Model Scalability Validation:} Experiments on a 70B-parameter model confirm that our findings generalize across scales, with optimizations yielding consistent relative gains despite absolute resource increases. However, we demonstrate that the effectiveness of certain techniques varies with model size, as optimization strategies successful for 8B models may not transfer effectively to larger architectures.

\end{itemize}

The rest of the paper is organized as follows. Section~\ref{methodology} details the model architecture, the evaluation metrics, the inference optimization algorithms selected for this study, and the experimental setup, including all hardware specifications. Sections~\ref{sec:lvl1} and~\ref{sec;lvl2} presents the insights obtained from applying each optimization individually and in combination under long-context workloads. Section~\ref{sec:70b} analyzes scalability, comparing the performance of larger models with that of smaller, architecturally identical counterparts. Section~\ref{sec:conclusion} concludes the paper by summarizing the primary findings.

\section{Notation and Background}

To clarify the terminology and notation used throughout this paper, we define several essential terms as follows. \textbf{Context} ($\mathcal{C}$) refers to the portion of the input providing necessary background information or support, represented as a sequence of tokens $\mathcal{C} = (t_1, t_2, \dots, t_n)$, where each $t_i$ is an individual token. The \textbf{Query} ($\mathbf{q}$) is a sequence of tokens directly requesting an action or output based on the given context. A \textbf{Prompt} ($\mathbf{p}$) combines context and query, formally defined as $\mathbf{p} = \mathcal{C} \parallel \mathbf{q}$, where "$\parallel$" is the concatenation operator, and the ordering of context and query is arbitrary. A \textbf{Compressed Prompt} ($\mathbf{p}_c$) similarly uses a compressed form of the context, denoted as $\mathbf{p}_c = \mathcal{C}_c \parallel \mathbf{q}$. Furthermore, the \textbf{Model} ($\mathcal{M}$) represents the large language model employed for inference, and its quantized and pruned variants are denoted as $\mathcal{M}_q$ and $\mathcal{M}_p$, respectively. Lastly, \textbf{Model Weights} ($W$) refer to parameters within a single transformer layer $l$ of the model. We use \enquote{quant} as short for quantization. \enquote{KV} short for Key Value vectors and c as prompt compression.

\section{Methodology}\label{methodology}

We present a systematic evaluation of assessing optimization techniques applied to newer architecture based large language models, \llamaModel with 8 billion parameters Mistral-Nemo with 12 billion parameters. We use \llamaModel with 70 billion parameters for our scalability analysis. Our approach employs a multi-dimensional analysis comprising both quality metrics like F1 score, precision, recall, and hallucination rates and efficiency metrics including throughput, latency, and memory consumption. This methodology enables us to establish correlations between optimization techniques and their impacts across model scales. 

\subsection{Model Selection}
Architecture of both \llamaModel and Mistral-Nemo supports an extended context size of up to 128K tokens. Selecting \llamaModel 8B allows comparison with its larger 70 billion parameter variant, as both models share the same context window capabilities and architecture, allowing us to examine how our findings scale across model sizes while maintaining consistent context handling characteristics.

\subsection{Algorithm Selection}
The following individual optimization techniques are denoted as level-1 algorithms when applied independently to the base model architecture. Table~\ref{tab:all_methods} defines the notation for the methods we used for evaluation.

\subsubsection{Pruning}
We utilize \textit{Minitron}~\cite{minitron2024} family of pruned models: a 4B \llamaModel 8B derivative retaining 50\% of parameters via width pruning of embeddings and MLP intermediates, and an 8B Mistral‑Nemo variant pruned from its 12B base to 66\%. The pruned undergoes retraining via distillation with Llama 3.1 405B as the teacher model. For scalability analysis comparison, we employed \textit{Nemotron}, a 51B-parameter obtained through identical pruning methodology from Llama-3.1 70B.

\subsubsection{Quantization}
We employ \textit{GPTQ}~\cite{frantar2022gptq} quantization to compress all model weights except the language modeling head (the final output layer that predicts the next token logits) to 4-bit integer precision, while maintaining FP16 activations throughout the network. We only used 4-bit quantization for optimized hardware performance as we use AutoGPTQ\footnote{\url{https://github.com/AutoGPTQ/AutoGPTQ}}, that utilizes Marlin kernel integration which provided optimal performance for for 4-bit integer weight and FP16 activation. For the 70B model, we utilized exllamav2\footnote{\url{https://github.com/turboderp-org/exllamav2}} to execute the experiments with pipeline parallelism. The quantization calibration was performed using the TriviaQA subset of the LongBench~\cite{bai-etal-2024-longbench}.

\subsubsection{Prompt Compression}
We used LLM-Lingua2~\cite{pan-etal-2024-llmlingua}, a BERT-based compression model, operating on CPU with a 2x compression rate to manage extended context lengths while avoiding GPU memory limitations. LLM-Lingua was selected for its minimal token-dropping overhead and demonstrated state-of-the-art prompt compression performance.

\subsubsection{KV Cache Compression}
KIVI~\cite{liu2024kivi} is adopted for KV-cache quantization to 4-bit integer width. The KV cache quantization used in this study  leverages HuggingFace's framework, drawing inspiration from the KIVI methodology.

\begin{figure*}[t]
    \caption{Performance Across Level-1 Optimization Methods for \llamaModel8B and Mistral-Nemo}
  \includegraphics[width=0.245\textwidth,height=4.1cm]{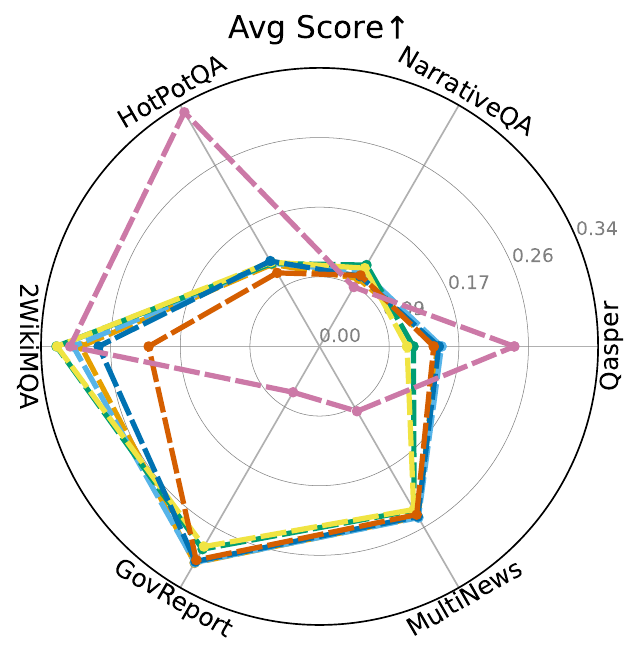}
  \includegraphics[width=0.245\textwidth,height=4.1cm]{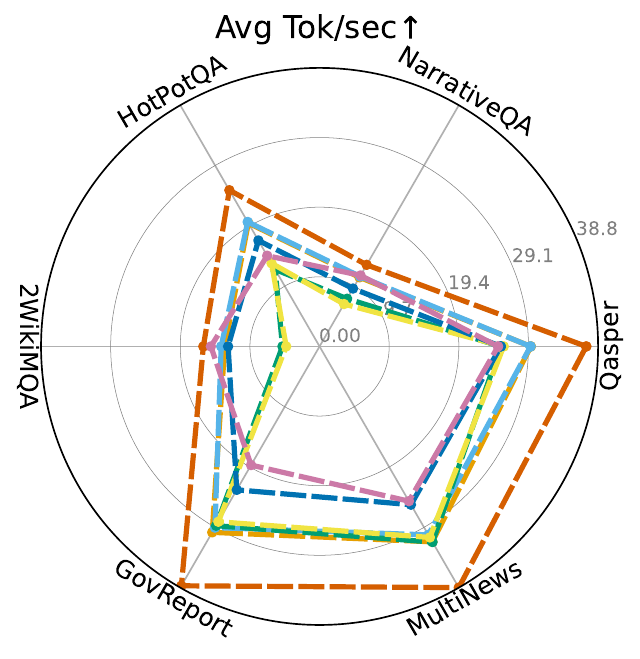}
  \includegraphics[width=0.245\textwidth,height=4.1cm]{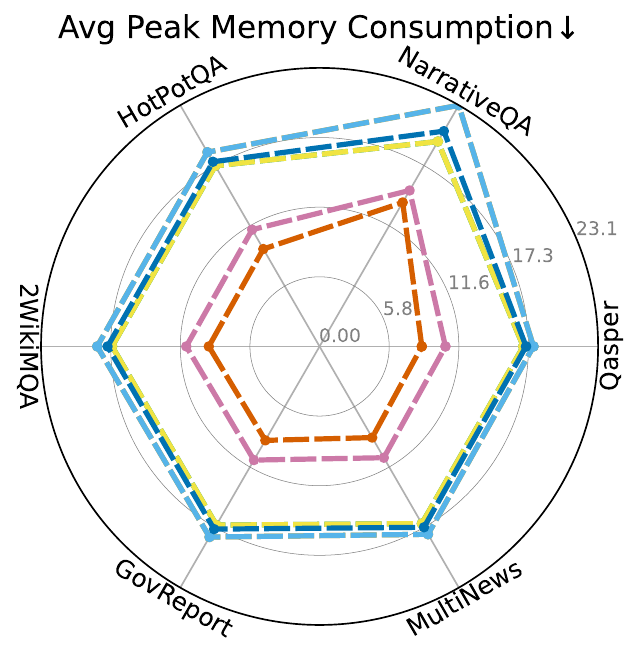}
  \includegraphics[width=0.245\textwidth,height=4.1cm]{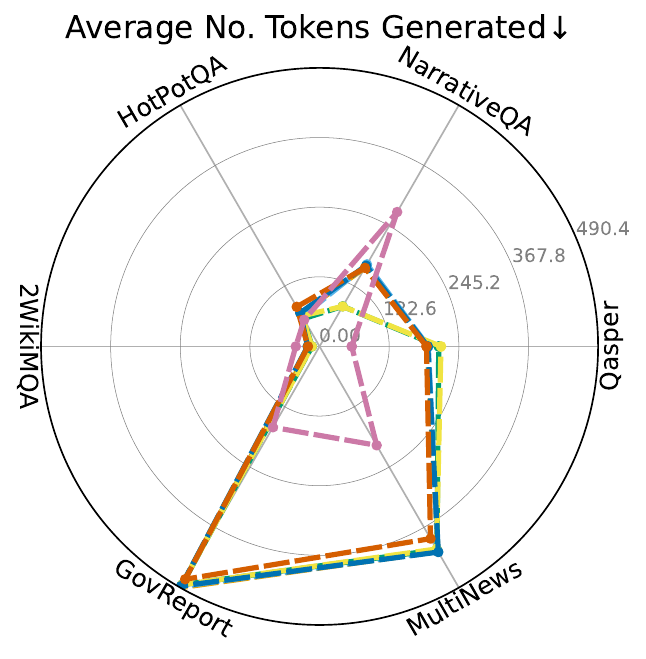}
  \includegraphics[width=0.245\textwidth,height=4.1cm]{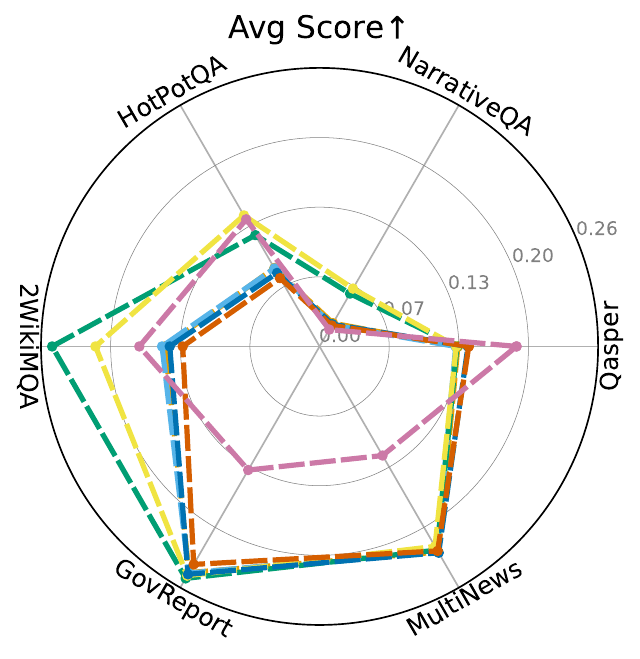}
  \includegraphics[width=0.245\textwidth,height=4.1cm]{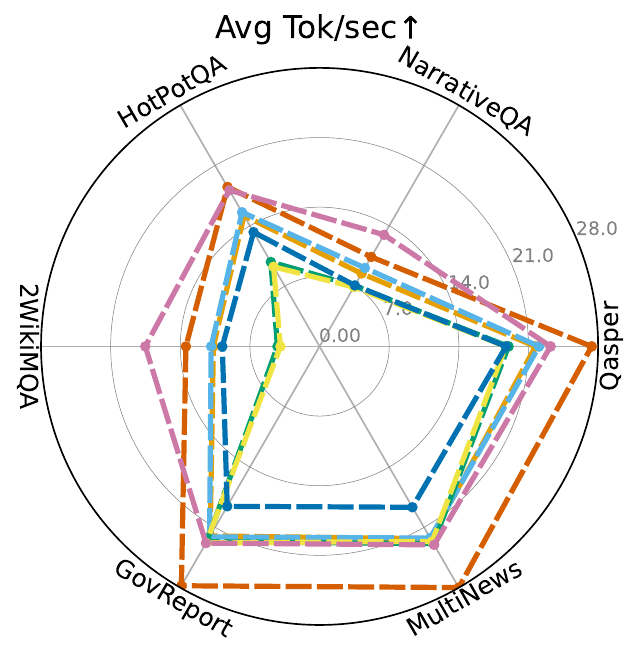}
  \includegraphics[width=0.245\textwidth,height=4.1cm]{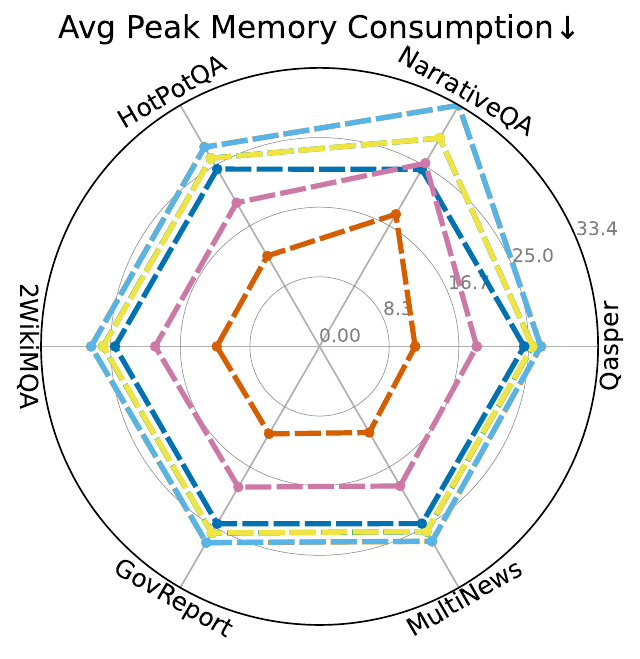}
  \includegraphics[width=0.245\textwidth,height=4.1cm]{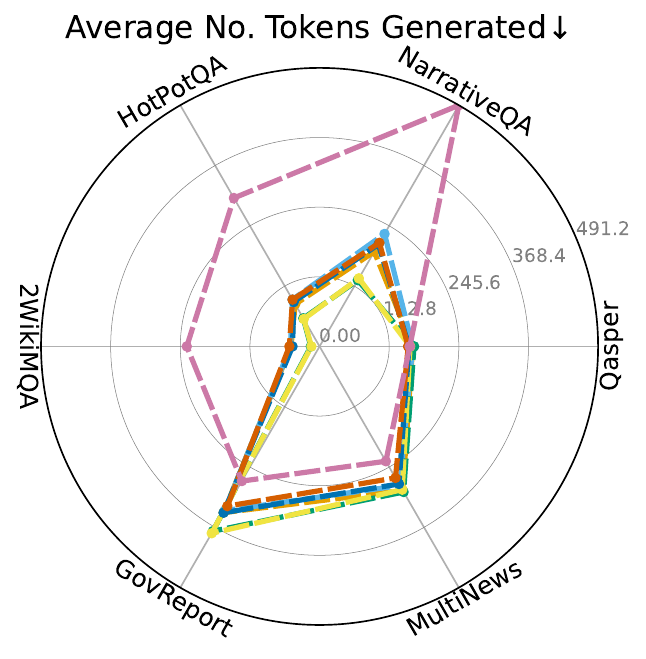}
  \centering
  \includegraphics[width=0.8\textwidth,height=0.6cm]{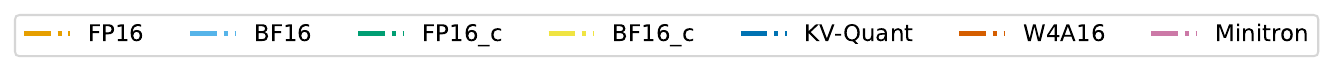}
  \label{fig:radar_all_lev1}
\end{figure*}

\subsubsection*{Multi-level Strategies} We also study the system performance of combined inference optimization algorithms denoted as level-2 algorithms, as combining these techniques  creates synergistic efficiency gains that exceed those achievable through individual methods alone. Let $\mathcal{M}$ be the original model architecture. The level-2 optimization techniques combine multiple level-1 algorithms to achieve compounded efficiency gains. For instance, applying pruning $\mathcal{P}$ followed by quantization $\mathcal{Q}$ yields $\mathcal{Q}(\mathcal{P}(\mathcal{M}))$, where Minitron pruning is combined with GPTQ 4-bit quantization. Similarly, integrating prompt compression via LLM-Lingua2 with KIVI's KV-cache compression allows for simultaneous optimization of both memory footprint and inference latency. These composite approaches enable multiplicative benefits while managing the accuracy-efficiency trade-offs of individual techniques. The order of these methods applied does matter. The order in which the optimization methods are applied is token dropping $\rightarrow$ pruning $\rightarrow$ quantization $\rightarrow$ KV cache quantization. This specific sequence ensures that each successive technique are compatible. 

\subsection{Dataset Selection}
In our study, we employ LongBench \cite{bai-etal-2024-longbench} to evaluate long-context LLM capabilities across six carefully selected datasets spanning three fundamental text processing tasks: i) Single document question answering or SDQA (NarrativeQA, Qasper), ii) Multi-document question answering or MDQA (HotpotQA, 2WikiMQA), and iii) summarization (GovReport, MultiNews). This selection enables assessment across diverse cognitive demands: single-document QA tests deep comprehension within extended narratives, multi-document QA evaluates information synthesis and cross-referencing across multiple sources, and summarization challenges models to distill extensive content into concise outputs. Despite the limited number of samples in each dataset (typically in the few hundreds), LongBench maintains consistent quality across these diverse tasks, making it suitable for our evaluation. This balanced approach provides insights into model performance across varying degrees of retrieval complexity and generation requirements, allowing for rigorous assessment of how different model optimizations affect long-context processing capabilities.

\begin{table}[t]
\footnotesize
\caption{Optimization techniques for LLM inference. Comp and quant indicates compression and  quantization respectively}
\label{tab:all_methods}
\begin{tabular}{|ll|}
\hline
\multicolumn{1}{|c|}{\textbf{Notation}} & \textbf{Optimization Technique} \\ \hline
\multicolumn{2}{|c|}{\textbf{Level 1}}                        \\ \hline
\multicolumn{1}{|l|}{FP16}     & Float point 16-bit precision                   \\ \hline
\multicolumn{1}{|l|}{BF16}     & Brain float point 16-bit precision             \\ \hline
\multicolumn{1}{|l|}{FP16\_c}  & FP16 with prompt compression        \\ \hline
\multicolumn{1}{|l|}{BF16\_c}  & BF16 with prompt compression        \\ \hline
\multicolumn{1}{|l|}{KV-Q} & Key-Value cache quantization                      \\ \hline
\multicolumn{1}{|l|}{W4A16}    & 4-bit weight, 16-bit activation quant      \\ \hline
\multicolumn{1}{|l|}{Minitron} & Pruned Model              \\ \hline
\multicolumn{2}{|c|}{\textbf{Level 2}}                           \\ \hline
\multicolumn{1}{|l|}{KV-Q+c}   & KV-Q with prompt compression  \\ \hline
\multicolumn{1}{|l|}{W4A16+c}  & W4A16 with prompt compression           \\ \hline
\multicolumn{1}{|l|}{Mini+c}   & Minitron with prompt compression                  \\ \hline
\multicolumn{1}{|l|}{W4A16 + KV\textsuperscript{Q}} & W4A16 with KV-Q \\ \hline
\multicolumn{1}{|l|}{Mini+KV\textsuperscript{Q}} & Minitron with KV-Q        \\ \hline
\multicolumn{1}{|l|}{Mini+W4A16} & Minitron with W4A16            \\ \hline
\multicolumn{2}{|c|}{\textbf{Scalability Analysis }}                           \\ \hline
\multicolumn{1}{|l|}{Nemotron} &  Pruned variant of \llamaModel 70B           \\ \hline
\end{tabular}
\vspace*{-\baselineskip}
\end{table}

\subsection{Metrics}
We use the \textit{QA-F1} as our primary scoring metric for both SDQA and MDQA, and \textit{ROUGE-L} for summarization evaluation. ROUGE-L is an F-measure based on the longest common subsequence (LCS), where precision and recall are computed using the LCS length between predicted and reference texts. The F1 score for QA tasks is defined as the harmonic mean of precision and recall, where precision is the fraction of predicted word occurrences matching the ground truth, and recall is the fraction of ground truth word occurrences found in the prediction.  We refer to these metrics as \textit{score} in our tables and figures. \textit{Hallucination score} (Hall. Score) is defined as $= \frac{|E_h|}{|E|}$ where $E$ is the set of extracted entities, noun chunks and tokens from answer and $E_h \subseteq  E$ set of hallucinated elements that were not found in context ($\C$). An element $e \in E$ is considered hallucinated if its fuzzy matching similarity score (FMSS) $< \epsilon,  \forall s \in \C$. Hall. Score $\in [0,1]$, where a lower value indicates fewer hallucinations and a higher value indicates that most extracted elements were not found in the source. FMSS is fuzzy string matching based on Levenshtein distance. We set $\epsilon = 0.8$ for all of our experiments.


\subsection{Experiment Setup}
All experiments were conducted on NVIDIA A100 GPUs with 40GB of VRAM, supported by 32GB of system memory and 16 CPU cores. This configuration was chosen to ensure consistent resource allocation across all experimental runs while maintaining sufficient computational capacity for handling long-context operations.

For our baseline models, we selected both FP16 and BF16 as this dual-baseline approach was implemented to evaluate the efficacy of BF16, which has demonstrated advantages in training scenarios, specifically in the context of long-sequence inference tasks. The A100 GPU architecture is proved capable of accommodating either model variant while supporting context lengths of up to 50,000 tokens in a single GPU configuration because of its ample on-board memory. We set the \texttt{batch\_size}=1 to ensure that the entire text fits within the available GPU memory. 
Table~\ref{tab:all_methods} shows all the optimization techniques that we used in our study. The experiment setup for scalability analysis is discussed in section~\ref{sec:impl_70b}.

\section{Experiments and Results}
For comprehensive evaluation of model performance and resource utilization, we collected multiple key metrics during our experiments. For each task, we measured the task-specific performance scores to assess output quality. We tracked the generation latency to understand computational efficiency, measuring the time taken from input processing to complete text generation. To quantify output characteristics, we recorded the number of tokens generated across different tasks and contexts. Additionally, we monitored peak memory consumption during text generation, providing insights into the memory requirements of different optimization techniques and their impact on resource utilization.
\subsection{Level 1 - Benchmarking Individual Inference Optimization Algorithms}\label{sec:lvl1}
Figure~\ref{fig:radar_all_lev1} presents the evaluation of individual optimization methods across three tasks. Each column corresponds to a task, which has two subtasks. Metrics for which higher values are better are indicated by~$\uparrow$, while metrics for which lower values are better are indicated by~$\downarrow$. The method that performs best in each subtask is highlighted in green, while the worst-performing methods are highlighted in red. The radar plot summarizing system evaluation across optimization methods is presented in Figure~\ref{fig:radar_all_lev1}. Based on our experiment, we made the following observations.

$\myCircled{1}$ \textbf{Weight quantization affords the greatest memory savings and excels at long‑context summarization}. The quantized \llamaModel and Mistral‑Nemo models cut GPU memory use by 1.97x and 2.17x and throughput is increased by 25\%, while ROUGE‑L score falls by only 1.3\% and 1.8\%. These gains arise because only the static weights are quantized; activations and freshly generated tokens remain full‑precision, preserving attention quality. The model performs particularly well on summarization tasks, where it consistently outperforms prompt compression methods, suggesting that quantization approaches may be more advantageous than token dropping strategies for long-context summarization. These findings align with the findings of~\cite{yuan-etal-2024-kv} for processing long contexts through LLM. Quantization reduces question‑answering accuracy by 13.5\%, driven by lower precision and higher hallucination score generated tokens appear in neither the gold answer nor the source context. Our analysis shows that quantization yields the greatest system‑level speed and memory gains, yet beyond summarization it degrades text quality, reducing average scores across six tasks by 9.45\% for \llamaModel and 3.17\% for Mistral‑Nemo.

$\myCircled{2}$ \textbf{Structured pruning with knowledge distillation markedly shrinks model size but introduces task‑specific trade‑offs}. In six long‑context QA benchmarks, pruned Minitron variants outperform their full‑size baselines, raising aggregate accuracy by 13.36\% for \llamaModel and 1.41\% for Mistral‑Nemo. The improvement is most pronounced in extractive QA as it result in 55.20\% increase in accuracy for \llamaModel and 25.13\% for Mistral-Nemo. Conversely, summarization quality collapses resulting in ‑70\% decrease for \llamaModel and ‑46\% for Mistral-Nemo, implying that pruning removes capacity essential for synthesizing coherent summaries. Although halving LLaMA‑Minitron's parameters cuts memory use by 1.65x, its throughput falls to 0.86x the baseline, demonstrating that aggressive pruning can reduce compute efficiency (see observation $\myCircled{9}$). Our text quality analysis as shown in Table~\ref{tab:avg_score_qa_lvl1} shows the pruned models' higher precision boosts their F1 on QA, but this same selectivity underscores the need for cautious, task‑aware pruning strategies.

\vspace*{-0.5em}
\begin{table}[h]
\centering
\caption{Average Output Quality of Level-1 Methods of \llamaModel8B across 4 long context QA tasks}
\label{tab:avg_score_qa_lvl1}
\resizebox{\columnwidth}{!}{
\begin{tabular}{|l|l|l|l|l|}
\toprule
    & F1 QA $\uparrow$ & Prec $\uparrow$ & Recall $\uparrow$ & H. Score $\downarrow$ \\
    \midrule
    FP16 & 16.19 & 12.02 & 54.00 & 36.43 \\
    BF16 & 16.85 & 12.50 & 54.68 & 36.41 \\
    FP16\_c & 16.72 & 13.13 & 50.45 & 32.91 \\
    BF16\_c & 16.45 & 12.88 & 51.67 & 33.17 \\
    KV-Q & 15.90 & 11.46 & \textbf{\cellcolor{blue!15}{55.92}} & \cellcolor{red!40}{37.40} \\
    W4A16 & \cellcolor{red!40}{13.83} & \cellcolor{red!40}{9.70} & 53.68 & 36.42 \\
    Minitron & \textbf{\cellcolor{blue!15}{23.95}} & \textbf{\cellcolor{blue!15}{23.04}} & \cellcolor{red!40}{36.08} & \textbf{\cellcolor{blue!15}{19.01}} \\
    \bottomrule
    \end{tabular}
}
\end{table}

\myCircled{3} \textbf{Minitron's design favors precision over comprehensive retrieval, yielding high accuracy and low hallucination rates but struggling with recall}. Table~\ref{tab:avg_score_qa_lvl1} shows that Minitron achieves the highest F1 QA score (23.95) and precision (23.04), but shows the lowest recall (36.08) and hallucination score (19.01), suggesting that it answers questions with high accuracy but retrieves significantly less relevant information from the context. Minitron's behavior stems from strategic parameter reduction: with only half the parameters of the base model, pruning eliminates uncertainty-generating connections while distillation from its 405B parameter teacher transfers only high-confidence knowledge. This creates a highly discriminative model that prioritizes accuracy over comprehensiveness, responding only when certain, resulting in exceptional precision and low hallucinations, but struggling to extract all relevant information from long documents due to its limited parameter capacity, thus exhibiting lower recall. 

\myCircled{4} \textbf{Textual prompt compression, KV quantization, and token‑dropping each deliver only marginal memory savings.} These savings ranges from 1.06x to 1.08x  while reducing throughput 0.70x to 0.85x increasing hallucination risk, and inconsistently altering token output, so their limited benefits such as KV quantization's slightly higher recall for or prompt compression’s niche MDQA gains rarely justify the added overhead, highlighting other methods as the more promising path to long‑context efficiency. KV quantization also produces the longest outputs, on average. The reduced precision KV representations appear to weaken the model’s ability to recognize semantic redundancy, prompting it to generate more tokens than necessary. In addition, KV quantization introduces per step latency: after each token is produced, its key and value tensors must be quantized before they are appended to the attention cache and used to compute the next attention scores. This overhead can offset some of the memory savings that the method is designed to provide especially in the long context as every token as KV for token has to quantized first before starting the token generation process.




\subsection{Level 2 - Evaluating Compounding Effect of Inference Optimization Algorithms}\label{sec;lvl2}
\begin{figure*}[t]
\centering
\caption{Comparative Performance of Level-2 Optimization Methods vs. FP16 Baseline}
\begin{tabular}{@{\hspace*{-0.4em}}cccc}
\includegraphics[width=0.23\textwidth]{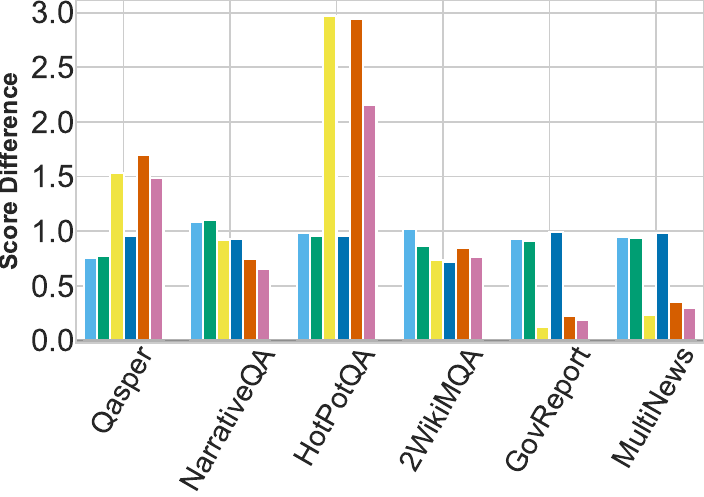} &
\includegraphics[width=0.23\textwidth]{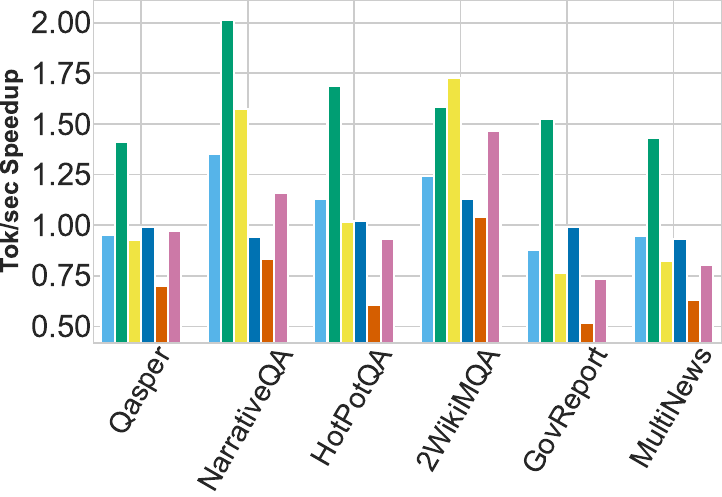} &
\includegraphics[width=0.23\textwidth]{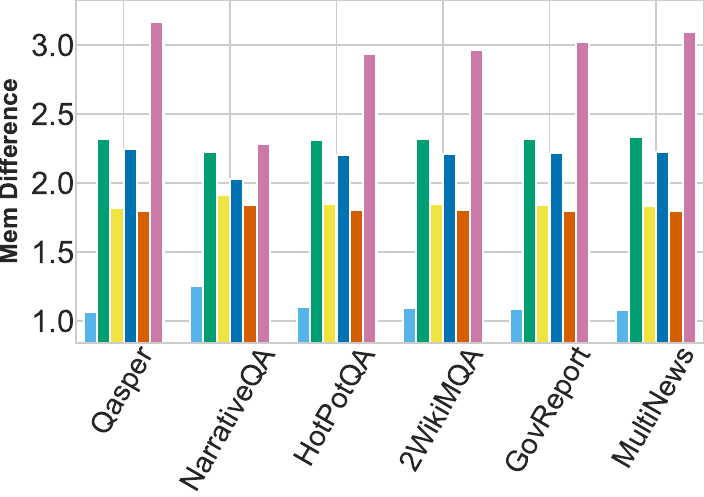} &
\includegraphics[width=0.23\textwidth]{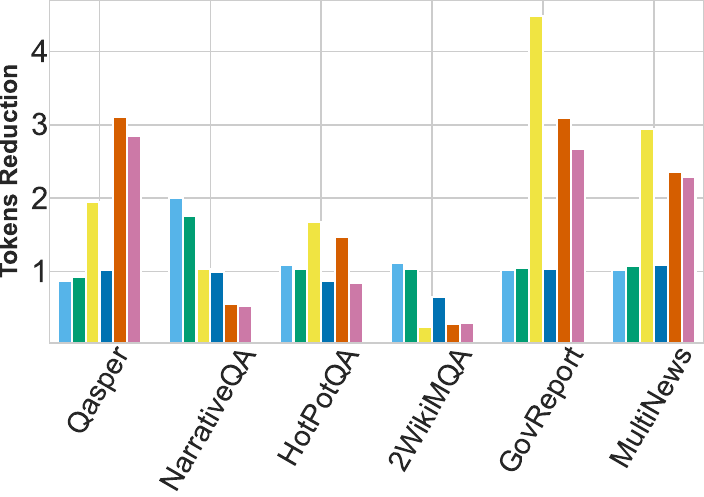}   \\
\includegraphics[width=0.23\textwidth]{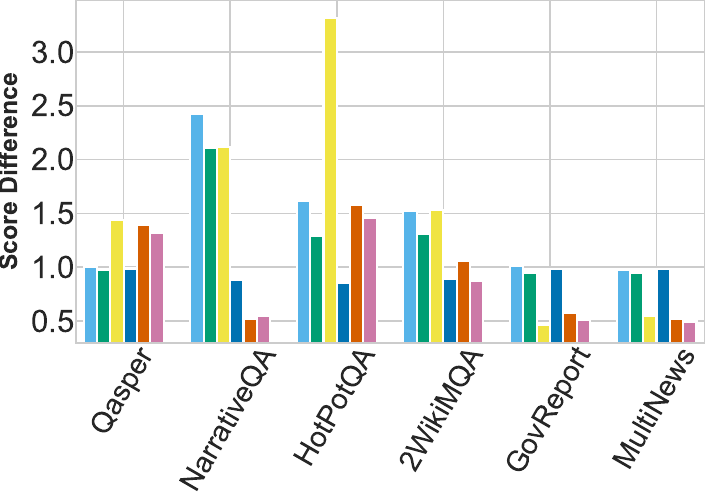} &
\includegraphics[width=0.23\textwidth]{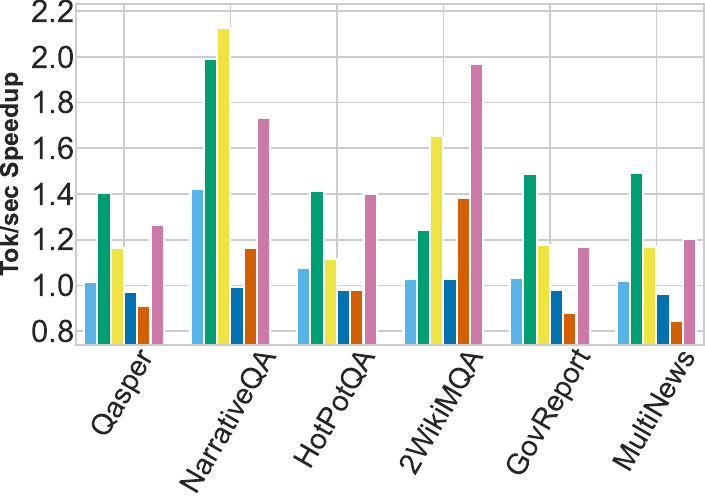} &
\includegraphics[width=0.23\textwidth]{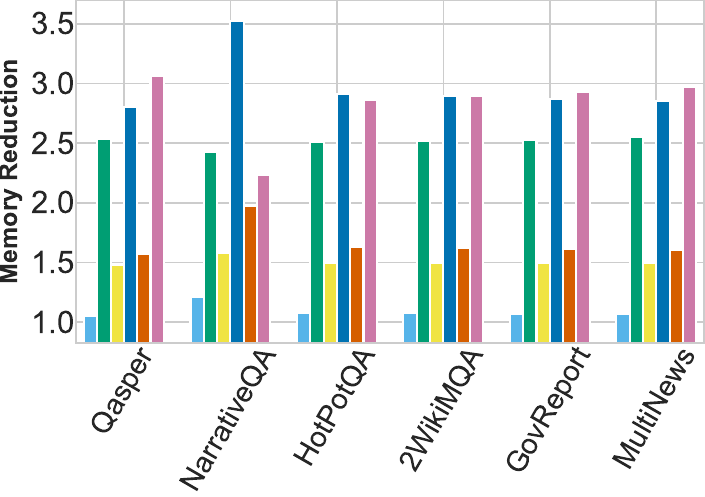} &
\includegraphics[width=0.23\textwidth]{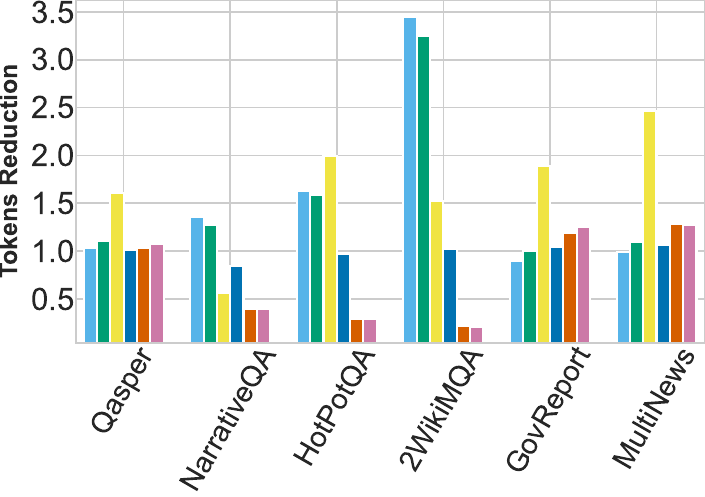}   \\
\multicolumn{4}{c}{
  \includegraphics[width=0.7\textwidth]{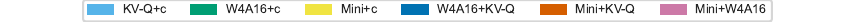}
} \\
\label{fig:all_perf_diff_lvl2}
\end{tabular}
\end{figure*}

Based on our initial analysis comparing FP16 and BF16 in the level-1 optimization, we selected FP16 for the two-level optimization experiments. This decision was informed by the marginal performance difference between the two formats, with BF16 showing only a slight advantage in accuracy (0.203 versus 0.199) while FP16 demonstrated marginally better inference time (9.73 versus 9.81 seconds). This selective approach also reduced the experimental complexity in level-two optimization by eliminating redundant format combinations. The performance gain across all tasks is shown in Figure~\ref{fig:all_perf_diff_lvl2}. It has the same structure as Table~\ref{tab:all_lvl2_metrics}. A radar plot summarizing system evaluation across level-2 optimization methods is shown in Figure~\ref{fig:radar_all_lvl2_all}. Analysis of our level-two optimization experiments reveal the following insights:

\myCircled{5} \textbf{Maximizing performance from combined inference optimizations requires complementary system‑level adjustments.} Pruning and quantization together cut memory consumption by up to 2.9x yet raise throughput by only 1.46x. Individually, each technique yields the top single‑pass speed‑up in our tests, but stacking them does not compound that benefit. Pairing methods whose strengths are complementary can produce larger gains. For instance, combining token dropping with weight quantization boosts throughput by 1.61x on Llama and 1.51x on Mistral well above the 1.25x ceiling achieved by any single optimization. The muted acceleration from the pruning quantization pair is largely an artifact of our setting batch size = 1 the extra GPU memory they free cannot translate into speed unless the batch size is increased. Merely stacking the highest‑performing algorithms does not automatically deliver optimal results.

\myCircled{6} \textbf{Identifying an inference‑optimization strategy that simultaneously maximizes accuracy, throughput, and memory efficiency is inherently task‑dependent and therefore non‑trivial.} Selecting a single \enquote{best} method is complicated by the fact that different tasks favor different combinations of optimizations. On average, weight pruning coupled with KV quantization yields the highest evaluation scores for \llamaModel, whereas pruning combined with token dropping performs best for Mistral‑NeMo. The KV‑quantization and token dropping step lowers throughput relative to other methods, yet it delivers a substantial memory saving that can be critical in constrained environments. 

\subsection{Model Scalability Analysis}\label{sec:70b}

\begin{table*}[t]
    \centering
    \caption{Evaluation Results on Model Scalability}
  \begin{tabular}{@{\hspace{-0.05cm}}m{5.0cm}m{5.0cm}m{5.0cm}}  
    \begin{minipage}[t][2.0cm][t]{0.34\textwidth}
    \centering
      \resizebox{\textwidth}{!}{  
\begin{tabular}{m{1.95cm}|>{\centering\arraybackslash}m{1.14cm}>{\centering\arraybackslash}m{1.11cm}>{\centering\arraybackslash}m{1.08cm}>{\centering\arraybackslash}m{1.53cm}}
                \toprule
                \textbf{Qasper}&  Score $\uparrow$ & Tok/s $\uparrow$  & Mem $\downarrow$ & No. Tok $\downarrow$ \\
                \midrule
FP16 & 0.201 & \textbf{\textcolor{red}{4.46}} & \textbf{\textcolor{red}{152.82}} & 136.17 \\
FP16\_c & 0.183 & 4.67 & 146.99 & 112.09 \\
KV-Q & 0.056 & 7.11 & 150.05 & 471.03 \\
W4A16 & \textbf{\textcolor{red}{0.055}} & \textbf{\textcolor{green}{20.18}} & \textbf{\textcolor{green}{61.79}} & \textbf{\textcolor{red}{473.02}} \\
Nemotron & \textbf{\textcolor{green}{0.257}} & 5.29 & 122.00 & \textbf{\textcolor{green}{81.75}} \\
                \bottomrule
            \end{tabular}
        }
        \end{minipage}
        &         
    \begin{minipage}[t][2.0cm][t]{0.34\textwidth}
        \centering
        \resizebox{\textwidth}{!}{  
\begin{tabular}{m{1.95cm}|>{\centering\arraybackslash}m{1.14cm}>{\centering\arraybackslash}m{1.11cm}>{\centering\arraybackslash}m{1.08cm}>{\centering\arraybackslash}m{1.53cm}}
            \toprule
            \textbf{HotpotQA}& Score $\uparrow$ & Tok/s $\uparrow$  & Mem $\downarrow$ & No. Tok $\downarrow$ \\
            \midrule
FP16 & 0.193 & \textbf{\textcolor{red}{2.83}} & \textbf{\textcolor{red}{159.04}} & 44.01 \\
FP16\_c & 0.262 & 2.94 & 149.75 & 32.61 \\
KV-Q & 0.022 & 6.15 & 150.04 & 428.29 \\
W4A16 & \textbf{\textcolor{red}{0.020}} & \textbf{\textcolor{green}{17.25}} & \textbf{\textcolor{green}{61.79}} & \textbf{\textcolor{red}{437.08}} \\
Nemotron & \textbf{\textcolor{green}{0.289}} & 3.19 & 130.77 & \textbf{\textcolor{green}{28.61}} \\
            \bottomrule
        \end{tabular}
        }
        \end{minipage}
        &
    \begin{minipage}[t][2.0cm][t]{0.34\textwidth}
        \centering
        \resizebox{\textwidth}{!}{  
\begin{tabular}{m{1.95cm}|>{\centering\arraybackslash}m{1.14cm}>{\centering\arraybackslash}m{1.11cm}>{\centering\arraybackslash}m{1.08cm}>{\centering\arraybackslash}m{1.53cm}}
            \toprule
            \textbf{Gov Report} & Score $\uparrow$ & Tok/s $\uparrow$  & Mem $\downarrow$ & No. Tok $\downarrow$ \\
            \midrule
FP16 & 0.301 & \textbf{\textcolor{red}{4.98}} & \textbf{\textcolor{red}{156.53}} & 454.61 \\
FP16\_c & \textbf{\textcolor{red}{0.281}} & 5.79 & 148.50 & \textbf{\textcolor{green}{442.72}} \\
KV-Q & 0.306 & 7.47 & 150.04 & 508.95 \\
W4A16 & \textbf{\textcolor{green}{0.312}} & \textbf{\textcolor{green}{20.78}} & \textbf{\textcolor{green}{61.79}} & \textbf{\textcolor{red}{509.21}} \\
Nemotron & 0.308 & 6.80 & 126.72 & 445.97 \\
            \bottomrule
            \end{tabular}
            }
            \end{minipage}
    \\
    \begin{minipage}[t][2.0cm][t]{0.34\textwidth}
        \centering
        \resizebox{\textwidth}{!}{  
\begin{tabular}{m{1.95cm}|>{\centering\arraybackslash}m{1.14cm}>{\centering\arraybackslash}m{1.11cm}>{\centering\arraybackslash}m{1.08cm}>{\centering\arraybackslash}m{1.53cm}}
                \toprule
                \textbf{NarrativeQA} & Score $\uparrow$ & Tok/s $\uparrow$  & Mem $\downarrow$ & No. Tok $\downarrow$ \\
                \midrule
FP16 & 0.133 & \textbf{\textcolor{red}{1.58}} & \textbf{\textcolor{red}{189.73}} & 137.32 \\
FP16\_c & \textbf{\textcolor{green}{0.193}} & 1.73 & 166.33 & \textbf{\textcolor{green}{44.37}} \\
KV-Q & \textbf{\textcolor{red}{0.019}} & 4.86 & 150.07 & 441.64 \\
W4A16 & 0.020 & \textbf{\textcolor{green}{12.34}} & \textbf{\textcolor{green}{62.26}} & \textbf{\textcolor{red}{442.25}} \\
Nemotron & 0.150 & 2.01 & 174.88 & 124.38 \\
                \bottomrule
            \end{tabular}
        }
        \end{minipage}
        &
            \begin{minipage}[t][2.0cm][t]{0.34\textwidth}
        \centering
        \resizebox{\textwidth}{!}{  
      \begin{tabular}{m{1.95cm}|>{\centering\arraybackslash}m{1.14cm}>{\centering\arraybackslash}m{1.11cm}>{\centering\arraybackslash}m{1.08cm}>{\centering\arraybackslash}m{1.53cm}}
             \toprule
        \textbf{2WikiMqa} & Score $\uparrow$ & Tok/s $\uparrow$  & Mem $\downarrow$ & No. Tok $\downarrow$ \\
            \midrule
FP16 & \textbf{\textcolor{green}{0.516}} & \textbf{\textcolor{red}{1.42}} & \textbf{\textcolor{red}{157.74}} & 10.18 \\
FP16\_c & 0.500 & 1.61 & 149.16 & \textbf{\textcolor{green}{9.49}} \\
KV-Q & \textbf{\textcolor{red}{0.029}} & 5.02 & 150.06 & 347.39 \\
W4A16 & 0.031 & \textbf{\textcolor{green}{14.22}} & \textbf{\textcolor{green}{61.79}} & \textbf{\textcolor{red}{355.77}} \\
Nemotron & 0.448 & 1.89 & 129.53 & 9.88 \\
            \bottomrule
            \end{tabular}
        }
        \end{minipage}
        &
    \begin{minipage}[t][2.0cm][t]{0.34\textwidth}
        \centering
        \resizebox{\textwidth}{!}{  
\begin{tabular}{m{1.95cm}|>{\centering\arraybackslash}m{1.14cm}>{\centering\arraybackslash}m{1.11cm}>{\centering\arraybackslash}m{1.08cm}>{\centering\arraybackslash}m{1.53cm}}
            \toprule
            \textbf{Multi News} & Score $\uparrow$ & Tok/s $\uparrow$  & Mem $\downarrow$ & No. Tok $\downarrow$ \\
            \midrule
FP16 & \textbf{\textcolor{green}{0.235}} & \textbf{\textcolor{red}{5.27}} & \textbf{\textcolor{red}{152.90}} & 378.97 \\
FP16\_c & \textbf{\textcolor{red}{0.227}} & 6.02 & 146.60 & \textbf{\textcolor{green}{358.51}} \\
KV-Q & 0.229 & 7.51 & 149.81 & 504.69 \\
W4A16 & 0.230 & \textbf{\textcolor{green}{21.16}} & \textbf{\textcolor{green}{61.78}} & \textbf{\textcolor{red}{505.22}} \\
Nemotron & 0.232 & 6.98 & 125.85 & 359.43 \\
            \bottomrule
        \end{tabular}
        }        
        \end{minipage}
        \end{tabular}

        \label{tab:all_stats_lvl1_70B}
 \end{table*}

\subsubsection*{Implementation Details for 70B}~\label{sec:impl_70b}
We set the number of CPU threads to be 16 with 64GB RAM for our experiments. We implemented pipeline parallelism to distribute the 70B parameter model across six A100 GPUs, each equipped with 40GB of VRAM. The GPUs were situated within a single computational cluster connected via NVLink. Due to our experimental configuration requiring a \texttt{batch\_size} of 1, pipeline parallelism was determined to be the optimal distribution strategy for this particular experimental setup. We ExLLama to distribute the quantized model accross GPUs. Our findings for scalability are: 

\myCircled{7} \textbf{Moderate pruning combined with knowledge distillation enables Nemotron to achieve strong performance but with constrained efficiency gains.} The pruned Nemotron model, having 27\% fewer parameters than the original 70B model, achieves superior performance, closely matching the text generation quality of the smaller Minitron model (with 50\% fewer parameters than the 8B baseline). Nemotron even surpasses baseline performance in one summarization task (0.232 vs. 0.235 baseline), illustrating that moderate pruning combined with knowledge distillation can exceed original model quality, whereas excessive pruning compromises complex task performance. However, Nemotron's efficiency improvements are proportionally limited: its throughput improvement (1.27x) aligns with parameter reduction, yet memory consumption is only reduced by 1.20x. This discrepancy likely arises from less aggressive pruning of attention mechanisms, resulting in larger KV-cache requirements. Additionally, Nemotron exhibits a 1.24x lower token generation rate, showing that parameter reductions proportionally affect inference speed but yield disproportionately lower memory savings as shown in Table~\ref{tab:all_stats_lvl1_70B}.

\myCircled{8} \textbf{Quantization transfers well from smaller to larger models but can incur significantly greater text quality degradation in 70B architectures, underscoring the need for careful optimization strategies.} 
While quantization provides similar advantages for the 70B model—such as higher throughput and lower memory consumption—its text quality suffers more acutely, dropping by 39\% compared to the 8B model's baseline. This more pronounced quality degradation arises from quantization errors accumulating across larger parameter spaces and may necessitate less aggressive approaches or post-quantization fine-tuning. Weight-only quantization generates the highest token counts in our experiments, with models often becoming repetitive after completing an answer degrading the quality; same with KV quantization. KV quantization in the 70B model reduces memory usage by 1.08x, slightly better than the 1.06x in the 8B model. 

\myCircled{9}\textbf{Maintaining power-of-2 dimensions in pruned Llama3.1 variants significantly enhances throughput compared to more aggressive pruning which reduces memory but compromises performance in certain tasks.} The pruned variants of Llama3.1, Nemotron (70B) and Minitron (8B), underwent identical distillation processes. Nemotron achieved a 1.26x increase in throughput and 1.2x reduction in memory consumption while retaining 73\% of the original parameters (a 27\% reduction). In contrast, Minitron experienced a 0.86x decrease in throughput. 
This disparity supports the hypothesis that dimension efficiency significantly affects performance: Nemotron maintains the original 70B model's power-of-2 hidden dimension (8192) and uses power-of-2 attention head groupings (8 to 64), while Minitron's non-power-of-2 hidden dimension (3072) yields suboptimal memory access patterns and kernel utilization on GPU hardware. Further evidence comes from the pruned Mistral‑NeMo variant: its power‑of‑two hidden dimension (4096) delivers higher throughput than its baseline, unlike the pruned Minitron version of \llamaModel8B. Consequently, even modest parameter reductions can provide notable throughput improvements when power-of-2 dimensions are preserved, whereas aggressively reducing parameters in non-power-of-2 dimensions may degrade computational efficiency despite greater memory savings.

These insights reveal that gains from inference optimizations are not simply additive. Future work could involve training meta‑routers that use real-time task classification to select the optimal inference technique or combination of techniques for each query, thereby maximizing system‑resource utilization while preserving the highest text quality.

\section{Conclusion}\label{sec:conclusion}
Our systematic analysis of optimization strategies for LLMs in long-context scenarios highlights key performance resource tradeoffs and model scalability insights. While pruning and quantization individually deliver meaningful resource efficiency, their naive combinations can significantly degrade text generation accuracy and quality due to compounded approximation errors. Moderate pruning combined with knowledge distillation, effectively maintains or surpasses original performance levels but yields proportionally limited efficiency improvements. Additionally, restricted pruning might be optimal for long context tasks, whereas aggressive parameter reduction compromises computational efficiency. Finally, optimization strategies that succeed in smaller (8B) models particularly quantization transfer effectively to larger (70B) models in terms of efficiency gains but result in disproportionately greater degradation in text quality. These findings underscore the need for efficient and synergetic large context-aware optimization approaches that balance accuracy, memory usage, and throughput, especially when scaling from smaller to larger model architectures.

\section*{Limitations}
All experiments were run with a batch size of 1. Increasing the batch size would require multi‑GPU execution, which introduces additional latencies, most notably the cost of transferring activations between devices. The magnitude of this overhead depends heavily on the hardware interconnect (e.g., PCIe vs. NVLink) and the specific topology, so our single‑GPU throughput results do not necessarily extrapolate to larger, distributed configurations.

We purposely limited the study to techniques with production‑ready GPU kernels, omitting methods for which only simulated or CPU‑bound implementations exist. While this focus ensures that the reported gains are achievable in practice especially for long‑context, interactive workloads it also means that some potentially valuable algorithms are not evaluated here.

The analysis is confined to system‑level metrics such as throughput, memory usage, and accuracy; it does not address content‑safety guardrails or alignment concerns. Practitioners should therefore apply appropriate safety measures e.g., content filtering or policy‑based refusals before deploying any models optimized according to these findings.

\iftrue
\section*{Acknowledgments}
This project is supported by the  Samsung Global Research Outreach Award and the U.S. Department of Energy, Advanced Scientific Computing Research (ASCR) of the U.S. Department of Energy under Contract No. DE-AC02-06CH11357.
\fi

\bibliography{my-bib}

\begin{thebibliography}{36}
\providecommand{\natexlab}[1]{#1}

\bibitem[{AI@Meta(2024)}]{llama3modelcard}
AI@Meta. 2024.
\newblock \href {https://github.com/meta-llama/llama3/blob/main/MODEL_CARD.md}
  {Llama 3 model card}.

\bibitem[{Ainslie et~al.(2023)Ainslie, Lee-Thorp, de~Jong, Zemlyanskiy,
  Lebr{\'o}n, and Sanghai}]{ainslie2023gqa}
Joshua Ainslie, James Lee-Thorp, Michiel de~Jong, Yury Zemlyanskiy, Federico
  Lebr{\'o}n, and Sumit Sanghai. 2023.
\newblock Gqa: Training generalized multi-query transformer models from
  multi-head checkpoints.
\newblock \emph{arXiv preprint arXiv:2305.13245}.

\bibitem[{Bai et~al.(2024)Bai, Lv, Zhang, Lyu, Tang, Huang, Du, Liu, Zeng, Hou,
  Dong, Tang, and Li}]{bai-etal-2024-longbench}
Yushi Bai, Xin Lv, Jiajie Zhang, Hongchang Lyu, Jiankai Tang, Zhidian Huang,
  Zhengxiao Du, Xiao Liu, Aohan Zeng, Lei Hou, Yuxiao Dong, Jie Tang, and
  Juanzi Li. 2024.
\newblock \href {https://doi.org/10.18653/v1/2024.acl-long.172} {{L}ong{B}ench:
  A bilingual, multitask benchmark for long context understanding}.
\newblock In \emph{Proceedings of the 62nd Annual Meeting of the Association
  for Computational Linguistics (Volume 1: Long Papers)}, pages 3119--3137,
  Bangkok, Thailand. Association for Computational Linguistics.

\bibitem[{Cappello()}]{cappelloauroragpt}
Franck Cappello.
\newblock Auroragpt: Exploring ai assistant for science.

\bibitem[{Chaturvedi et~al.(2024)Chaturvedi, Nichols, Singh, and
  Bhatele}]{chaturvedi2024hpc}
Aman Chaturvedi, Daniel Nichols, Siddharth Singh, and Abhinav Bhatele. 2024.
\newblock Hpc-coder-v2: Studying code llms across low-resource parallel
  languages.
\newblock \emph{arXiv preprint arXiv:2412.15178}.

\bibitem[{Chee et~al.(2023)Chee, Cai, Kuleshov, and
  De~Sa}]{NEURIPS2023_0df38cd1}
Jerry Chee, Yaohui Cai, Volodymyr Kuleshov, and Christopher~M De~Sa. 2023.
\newblock \href
  {https://proceedings.neurips.cc/paper_files/paper/2023/file/0df38cd13520747e1e64e5b123a78ef8-Paper-Conference.pdf}
  {Quip: 2-bit quantization of large language models with guarantees}.
\newblock In \emph{Advances in Neural Information Processing Systems},
  volume~36, pages 4396--4429. Curran Associates, Inc.

\bibitem[{Dao et~al.(2022)Dao, Fu, Ermon, Rudra, and
  R{\'e}}]{dao2022flashattention}
Tri Dao, Dan Fu, Stefano Ermon, Atri Rudra, and Christopher R{\'e}. 2022.
\newblock Flashattention: Fast and memory-efficient exact attention with
  io-awareness.
\newblock \emph{Advances in neural information processing systems},
  35:16344--16359.

\bibitem[{Dettmers et~al.(2022)Dettmers, Lewis, Belkada, and
  Zettlemoyer}]{dettmers2022llmint8}
Tim Dettmers, Mike Lewis, Younes Belkada, and Luke Zettlemoyer. 2022.
\newblock Gpt3. int8 (): 8-bit matrix multiplication for transformers at scale.
\newblock \emph{Advances in neural information processing systems},
  35:30318--30332.

\bibitem[{Dettmers et~al.(2023)Dettmers, Svirschevski, Egiazarian, Kuznedelev,
  Frantar, Ashkboos, Borzunov, Hoefler, and Alistarh}]{dettmers2023spqr}
Tim Dettmers, Ruslan Svirschevski, Vage Egiazarian, Denis Kuznedelev, Elias
  Frantar, Saleh Ashkboos, Alexander Borzunov, Torsten Hoefler, and Dan
  Alistarh. 2023.
\newblock Spqr: A sparse-quantized representation for near-lossless llm weight
  compression.
\newblock \emph{arXiv preprint arXiv:2306.03078}.

\bibitem[{Frantar et~al.(2022)Frantar, Ashkboos, Hoefler, and
  Alistarh}]{frantar2022gptq}
Elias Frantar, Saleh Ashkboos, Torsten Hoefler, and Dan Alistarh. 2022.
\newblock Gptq: Accurate post-training quantization for generative pre-trained
  transformers.
\newblock \emph{arXiv preprint arXiv:2210.17323}.

\bibitem[{Frantar et~al.(2024)Frantar, Castro, Chen, Hoefler, and
  Alistarh}]{frantar2024marlin}
Elias Frantar, Roberto~L Castro, Jiale Chen, Torsten Hoefler, and Dan Alistarh.
  2024.
\newblock Marlin: Mixed-precision auto-regressive parallel inference on large
  language models.
\newblock \emph{arXiv preprint arXiv:2408.11743}.

\bibitem[{Grattafiori et~al.(2024)Grattafiori, Dubey, Jauhri, Pandey, Kadian,
  Al-Dahle, Letman, Mathur, Schelten, Vaughan et~al.}]{grattafiori2024llama}
Aaron Grattafiori, Abhimanyu Dubey, Abhinav Jauhri, Abhinav Pandey, Abhishek
  Kadian, Ahmad Al-Dahle, Aiesha Letman, Akhil Mathur, Alan Schelten, Alex
  Vaughan, and 1 others. 2024.
\newblock The llama 3 herd of models.
\newblock \emph{arXiv preprint arXiv:2407.21783}.

\bibitem[{Hooper et~al.(2024)Hooper, Kim, Mohammadzadeh, Mahoney, Shao,
  Keutzer, and Gholami}]{NEURIPS2024_028fcbcf}
Coleman Hooper, Sehoon Kim, Hiva Mohammadzadeh, Michael~W. Mahoney,
  Yakun~Sophia Shao, Kurt Keutzer, and Amir Gholami. 2024.
\newblock \href
  {https://proceedings.neurips.cc/paper_files/paper/2024/file/028fcbcf85435d39a40c4d61b42c99a4-Paper-Conference.pdf}
  {Kvquant: Towards 10 million context length llm inference with kv cache
  quantization}.
\newblock In \emph{Advances in Neural Information Processing Systems},
  volume~37, pages 1270--1303. Curran Associates, Inc.

\bibitem[{Hu et~al.(2022)Hu, Shen, Wallis, Allen-Zhu, Li, Wang, Wang, Chen
  et~al.}]{hu2022lora}
Edward~J Hu, Yelong Shen, Phillip Wallis, Zeyuan Allen-Zhu, Yuanzhi Li, Shean
  Wang, Lu~Wang, Weizhu Chen, and 1 others. 2022.
\newblock Lora: Low-rank adaptation of large language models.
\newblock \emph{ICLR}, 1(2):3.

\bibitem[{Jiang et~al.(2023)Jiang, Wu, Lin, Yang, and Qiu}]{llmlingua23}
Huiqiang Jiang, Qianhui Wu, Chin-Yew Lin, Yuqing Yang, and Lili Qiu. 2023.
\newblock \href {https://doi.org/10.18653/v1/2023.emnlp-main.825} {{LLML}ingua:
  Compressing prompts for accelerated inference of large language models}.
\newblock In \emph{Proceedings of the 2023 Conference on Empirical Methods in
  Natural Language Processing}, pages 13358--13376, Singapore. Association for
  Computational Linguistics.

\bibitem[{Jin et~al.(2024)Jin, Du, Huang, Liu, Luan, Wang, and
  Xiong}]{jin-etal-2024-comprehensive}
Renren Jin, Jiangcun Du, Wuwei Huang, Wei Liu, Jian Luan, Bin Wang, and Deyi
  Xiong. 2024.
\newblock \href {https://doi.org/10.18653/v1/2024.findings-acl.726} {A
  comprehensive evaluation of quantization strategies for large language
  models}.
\newblock In \emph{Findings of the Association for Computational Linguistics:
  ACL 2024}, pages 12186--12215, Bangkok, Thailand. Association for
  Computational Linguistics.

\bibitem[{Jo et~al.(2024)Jo, Kim, Kim, and Kim}]{NEURIPS2024_f89221ed}
Dongwon Jo, Taesu Kim, Yulhwa Kim, and Jae-Joon Kim. 2024.
\newblock \href
  {https://proceedings.neurips.cc/paper_files/paper/2024/file/f89221edad5a6a4a54fcf247cb37cd62-Paper-Conference.pdf}
  {Mixture of scales: Memory-efficient token-adaptive binarization for large
  language models}.
\newblock In \emph{Advances in Neural Information Processing Systems},
  volume~37, pages 137474--137494. Curran Associates, Inc.

\bibitem[{Kurtic et~al.(2022)Kurtic, Campos, Nguyen, Frantar, Kurtz, Fineran,
  Goin, and Alistarh}]{kurtic-etal-2022-optimal}
Eldar Kurtic, Daniel Campos, Tuan Nguyen, Elias Frantar, Mark Kurtz, Benjamin
  Fineran, Michael Goin, and Dan Alistarh. 2022.
\newblock \href {https://doi.org/10.18653/v1/2022.emnlp-main.279} {The optimal
  {BERT} surgeon: Scalable and accurate second-order pruning for large language
  models}.
\newblock In \emph{Proceedings of the 2022 Conference on Empirical Methods in
  Natural Language Processing}, pages 4163--4181, Abu Dhabi, United Arab
  Emirates. Association for Computational Linguistics.

\bibitem[{LeCun et~al.(1989)LeCun, Denker, and Solla}]{lecun1989optimal}
Yann LeCun, John Denker, and Sara Solla. 1989.
\newblock Optimal brain damage.
\newblock \emph{Advances in neural information processing systems}, 2.

\bibitem[{Lee et~al.(2024)Lee, Park, Kwon, Oh, and
  Kwon}]{lee2024comprehensiveevaluationquantizedinstructiontuned}
Jemin Lee, Sihyeong Park, Jinse Kwon, Jihun Oh, and Yongin Kwon. 2024.
\newblock \href {https://arxiv.org/abs/2409.11055} {A comprehensive evaluation
  of quantized instruction-tuned large language models: An experimental
  analysis up to 405b}.
\newblock \emph{Preprint}, arXiv:2409.11055.

\bibitem[{Li et~al.(2024)Li, Ning, Wang, Liu, Shi, Yan, Dai, Yang, and
  Wang}]{li2024evaluatingquantizedlargelanguage}
Shiyao Li, Xuefei Ning, Luning Wang, Tengxuan Liu, Xiangsheng Shi, Shengen Yan,
  Guohao Dai, Huazhong Yang, and Yu~Wang. 2024.
\newblock \href {https://arxiv.org/abs/2402.18158} {Evaluating quantized large
  language models}.
\newblock \emph{Preprint}, arXiv:2402.18158.

\bibitem[{Li et~al.(2023)Li, Dong, Guerin, and Lin}]{selectivecontext23}
Yucheng Li, Bo~Dong, Frank Guerin, and Chenghua Lin. 2023.
\newblock \href {https://doi.org/10.18653/v1/2023.emnlp-main.391} {Compressing
  context to enhance inference efficiency of large language models}.
\newblock In \emph{Proceedings of the 2023 Conference on Empirical Methods in
  Natural Language Processing}, pages 6342--6353, Singapore. Association for
  Computational Linguistics.

\bibitem[{Lin et~al.(2024{\natexlab{a}})Lin, Tang, Tang, Yang, Chen, Wang,
  Xiao, Dang, Gan, and Han}]{lin2024awq}
Ji~Lin, Jiaming Tang, Haotian Tang, Shang Yang, Wei-Ming Chen, Wei-Chen Wang,
  Guangxuan Xiao, Xingyu Dang, Chuang Gan, and Song Han. 2024{\natexlab{a}}.
\newblock Awq: Activation-aware weight quantization for on-device llm
  compression and acceleration.
\newblock \emph{Proceedings of Machine Learning and Systems}, 6:87--100.

\bibitem[{Lin et~al.(2024{\natexlab{b}})Lin, Tang, Tang, Yang, Chen, Wang,
  Xiao, Dang, Gan, and Han}]{xiao2023awq}
Ji~Lin, Jiaming Tang, Haotian Tang, Shang Yang, Wei-Ming Chen, Wei-Chen Wang,
  Guangxuan Xiao, Xingyu Dang, Chuang Gan, and Song Han. 2024{\natexlab{b}}.
\newblock Awq: Activation-aware weight quantization for on-device llm
  compression and acceleration.
\newblock \emph{Proceedings of Machine Learning and Systems}, 6:87--100.

\bibitem[{Liu et~al.(2024)Liu, Yuan, Jin, Zhong, Xu, Braverman, Chen, and
  Hu}]{liu2024kivi}
Zirui Liu, Jiayi Yuan, Hongye Jin, Shaochen Zhong, Zhaozhuo Xu, Vladimir
  Braverman, Beidi Chen, and Xia Hu. 2024.
\newblock Kivi: A tuning-free asymmetric 2bit quantization for kv cache.
\newblock \emph{arXiv preprint arXiv:2402.02750}.

\bibitem[{Mahdavi et~al.(2023)Mahdavi, Liao, and
  Thrampoulidis}]{mahdavi2023memorization}
Sadegh Mahdavi, Renjie Liao, and Christos Thrampoulidis. 2023.
\newblock Memorization capacity of multi-head attention in transformers.
\newblock \emph{arXiv preprint arXiv:2306.02010}.

\bibitem[{Muralidharan et~al.(2024)Muralidharan, Turuvekere~Sreenivas, Joshi,
  Chochowski, Patwary, Shoeybi, Catanzaro, Kautz, and Molchanov}]{minitron2024}
Saurav Muralidharan, Sharath Turuvekere~Sreenivas, Raviraj Joshi, Marcin
  Chochowski, Mostofa Patwary, Mohammad Shoeybi, Bryan Catanzaro, Jan Kautz,
  and Pavlo Molchanov. 2024.
\newblock \href
  {https://proceedings.neurips.cc/paper_files/paper/2024/file/4822991365c962105b1b95b1107d30e5-Paper-Conference.pdf}
  {Compact language models via pruning and knowledge distillation}.
\newblock In \emph{Advances in Neural Information Processing Systems},
  volume~37, pages 41076--41102. Curran Associates, Inc.

\bibitem[{Pan et~al.(2024)Pan, Wu, Jiang, Xia, Luo, Zhang, Lin, R{\"u}hle,
  Yang, Lin, Zhao, Qiu, and Zhang}]{pan-etal-2024-llmlingua}
Zhuoshi Pan, Qianhui Wu, Huiqiang Jiang, Menglin Xia, Xufang Luo, Jue Zhang,
  Qingwei Lin, Victor R{\"u}hle, Yuqing Yang, Chin-Yew Lin, H.~Vicky Zhao, Lili
  Qiu, and Dongmei Zhang. 2024.
\newblock \href {https://doi.org/10.18653/v1/2024.findings-acl.57}
  {{LLML}ingua-2: Data distillation for efficient and faithful task-agnostic
  prompt compression}.
\newblock In \emph{Findings of the Association for Computational Linguistics:
  ACL 2024}, pages 963--981, Bangkok, Thailand. Association for Computational
  Linguistics.

\bibitem[{Sheng et~al.(2023)Sheng, Zheng, Yuan, Li, Ryabinin, Chen, Liang,
  R{\'e}, Stoica, and Zhang}]{chen2023flexgen}
Ying Sheng, Lianmin Zheng, Binhang Yuan, Zhuohan Li, Max Ryabinin, Beidi Chen,
  Percy Liang, Christopher R{\'e}, Ion Stoica, and Ce~Zhang. 2023.
\newblock Flexgen: High-throughput generative inference of large language
  models with a single gpu.
\newblock In \emph{International Conference on Machine Learning}, pages
  31094--31116. PMLR.

\bibitem[{Sreenivas et~al.(2024)Sreenivas, Muralidharan, Joshi, Chochowski,
  Patwary, Shoeybi, Catanzaro, Kautz, and Molchanov}]{sreenivas2024llm}
Sharath~Turuvekere Sreenivas, Saurav Muralidharan, Raviraj Joshi, Marcin
  Chochowski, Mostofa Patwary, Mohammad Shoeybi, Bryan Catanzaro, Jan Kautz,
  and Pavlo Molchanov. 2024.
\newblock Llm pruning and distillation in practice: The minitron approach.
\newblock \emph{arXiv preprint arXiv:2408.11796}.

\bibitem[{Sun et~al.(2024)Sun, Liu, Bair, and Kolter}]{sun2024a}
Mingjie Sun, Zhuang Liu, Anna Bair, and J~Zico Kolter. 2024.
\newblock \href {https://openreview.net/forum?id=PxoFut3dWW} {A simple and
  effective pruning approach for large language models}.
\newblock In \emph{The Twelfth International Conference on Learning
  Representations}.

\bibitem[{Touvron et~al.(2023)Touvron, Martin, Stone, Albert, Almahairi,
  Babaei, Bashlykov, Batra, Bhargava, Bhosale et~al.}]{touvron2023llama2}
Hugo Touvron, Louis Martin, Kevin Stone, Peter Albert, Amjad Almahairi, Yasmine
  Babaei, Nikolay Bashlykov, Soumya Batra, Prajjwal Bhargava, Shruti Bhosale,
  and 1 others. 2023.
\newblock Llama 2: Open foundation and fine-tuned chat models.
\newblock \emph{arXiv preprint arXiv:2307.09288}.

\bibitem[{Wang et~al.(2019)Wang, Grosse, Fidler, and
  Zhang}]{wang2019eigendamage}
Chaoqi Wang, Roger Grosse, Sanja Fidler, and Guodong Zhang. 2019.
\newblock Eigendamage: Structured pruning in the kronecker-factored eigenbasis.
\newblock In \emph{International conference on machine learning}, pages
  6566--6575. PMLR.

\bibitem[{Xiao et~al.(2023)Xiao, Lin, Seznec, Wu, Demouth, and
  Han}]{xiao2023smoothquant}
Guangxuan Xiao, Ji~Lin, Mickael Seznec, Hao Wu, Julien Demouth, and Song Han.
  2023.
\newblock {S}mooth{Q}uant: Accurate and efficient post-training quantization
  for large language models.
\newblock In \emph{Proceedings of the 40th International Conference on Machine
  Learning}.

\bibitem[{Yuan et~al.(2024)Yuan, Liu, Zhong, Chuang, Li, Wang, Le, Jin,
  Chaudhary, Xu, Liu, and Hu}]{yuan-etal-2024-kv}
Jiayi Yuan, Hongyi Liu, Shaochen Zhong, Yu-Neng Chuang, Songchen Li, Guanchu
  Wang, Duy Le, Hongye Jin, Vipin Chaudhary, Zhaozhuo Xu, Zirui Liu, and Xia
  Hu. 2024.
\newblock \href {https://doi.org/10.18653/v1/2024.findings-emnlp.266} {{KV}
  cache compression, but what must we give in return? a comprehensive benchmark
  of long context capable approaches}.
\newblock In \emph{Findings of the Association for Computational Linguistics:
  EMNLP 2024}, pages 4623--4648, Miami, Florida, USA. Association for
  Computational Linguistics.

\bibitem[{Zhang et~al.(2022)Zhang, Roller, Goyal, Artetxe, Chen, Chen, Dewan,
  Diab, Li, Lin et~al.}]{zhang2022opt}
Susan Zhang, Stephen Roller, Naman Goyal, Mikel Artetxe, Moya Chen, Shuohui
  Chen, Christopher Dewan, Mona Diab, Xian Li, Xi~Victoria Lin, and 1 others.
  2022.
\newblock Opt: Open pre-trained transformer language models.
\newblock \emph{arXiv preprint arXiv:2205.01068}.

\end{thebibliography}

\clearpage
\appendix

\section{Inference Optimization Algorithms and Background}
\subsubsection*{\textbf{Quantization}}
Quantization maps numbers from higher to lower-precision representations by reducing bit width (e.g., 32-bit floating-points
to 8-bit integers). For a value $\X$ with bit width $N$, integer uniform quantization is defined as $ \X_{\mathrm{INT}}=\left\lceil\frac{\X_{\mathrm{FP16}}-Z}{S}\right\rfloor $ 
where $S$ and $Z$ are the scaling factor and zero-point respectively. Two common scaling factor formulations are $
S_1=\frac{\max \left(\X_{\mathrm{FP16}}\right)-\min \left(\X_{\mathrm{FP16}}\right)}{2^{N-1}-1}$, $S_2=\frac{\max (|\X_{\mathrm{FP16}}|)}{2^{N-1}-1}$ where $S_1$ is the range used for asymmetric quantization, while $S_2$ is based on absolute maximum used in symmetric quantization.

For linear operation $Y = X \cdot W_i, $ where $\ W_i \in W$ is the activations and $X$ represents either input embeddings $\p_e$ or previous layer outputs. The quantization types for $\M$ are \textit{Weight-only Quantization}: Only $W$ is quantized while keeping activations in full precision. \textit{Activation-only Quantization}: Only activations $Y$ are quantized while keeping $W$ in full precision. \textit{Weight-Activation Quantization}: Both $W$ and $Y$ are quantized to lower precision.

\subsubsection*{Pruning}

Pruning is a process that involves the removal of components from the set $W$ or within its individual elements. 
This can be achieved either by removing the component or by setting certain entries of $W_i$ to zero~\cite{lecun1989optimal}. 
The candidate components to be pruned can be selected based on their magnitude~\cite{sun2024a} or their $n$-th order importance~\cite{kurtic-etal-2022-optimal, wang2019eigendamage}. Pruning can be either structured, un-structured or mixture of both. In un-structured pruning, individual weight entries in $W_i$ are set to zero, creating an irregular sparse representation that requires a minimum threshold of $k\%$ zero entries to achieve meaningful hardware performance gains. Conversely, structured pruning takes a more aggressive approach by eliminating entire structural components such as rows, columns, attention heads, or layers from the model $\M$. In our study, we used models pruned through structural pruning as it yields better hardware efficiency. 

\subsubsection*{Prompt Compression}
 In addition to traditional methods that mainly focus on optimizing large language models (LLMs) themselves, alternative techniques have been developed that operate directly on the input data. These techniques, known as context compression methods~\cite{selectivecontext23, llmlingua23} or token dropping, utilize a smaller LLM to eliminate non-essential words from the input. By reducing the number of tokens of input data in this manner, memory consumption is indirectly minimized, as the model is required to compute attention values only for the retained, significant words. Compressed Context ($\Cc$): Refers to a modified version of the input $\C$. $\Cc \subset \C$ which implies $|\Cc| < |\C|$.  The pruned input $\Cc$ is defined as $ \Cc = ( t_i \in C \mid \text{criterion}(t_i, \C)) $ where $ \text{criterion}(t_i, \C) $ is a function that determines whether $ t_i $ is to be kept for the given $\C$. The order of words in the pruned input $ \Cc $ is maintained.

\subsubsection*{KV-Cache Quantization}
For each token $t\in \mathbf{p}$, the key and value vectors for a single attention layer $l \in M$ are $K_t = h_t \, W_K$ and $V_t= h_t \, W_V \in \mathbb{R}^{d_k}$, respectively, and $h_t \in \mathbb{R}^{D}$ is the hidden state at layer $l$ for token $t$. Typically, the model dimension $D$ is split into multiple attention heads, each of dimension $d_k$, but we omit that detail here for clarity. Key and value for all tokens are cached to avoid recomputation. The KV-cache for the prompt is $K,V \in \mathbb{R}^{L \times T \times d_k}$ where both usually have the same dimensions. The quantization is applied to both $K \text{ and }V$, and thus, KV-cache quantization incurs additional computational overhead by quantizing each newly generated token requiring quantization operations at every autoregressive step.

\subsubsection*{Memory Growth For Long Context}
The total memory required during text generation is the sum of memory required to hold
model parameters $M_{\text{p}} $, intermediate computations within each transformer layer $ M_{\text{acts}} $, model's KV-cache $ M_{\text{KV}} $ and output logits $ M_{\text{logit}}$ . The total memory $M_{\text{total}}$ required to generate $N$ new tokens is equal to:
\begin{multline*}
\underbrace{M_{\text{p}} \times p_m}_{\text{Model}}
+\underbrace{ \left( L \times (N+C) \times H  \right) \times  p_a }_{M_{\text{acts}}} + \\
\underbrace{\left( 2 \times L \times \left( \frac{(N+C) \times H}{G} \right) \right)  \times p_k}_{M_{\text{KV}}}+ \\
\underbrace{\left(H \times |V|\right)\times p_m}_{M_\text{logit}}
\end{multline*}
Where $p_m$, $p_a$, and $p_k$ denote the bit precision for model's parameter, activation, and KV-cache, respectively. $L$ is the number of layers, $H= d_k \times h_q$ is the hidden dimension of each layer where $d_k$ is the head dimension and $h_q$ is the number of attention heads. $|V|$ denoted the vocabulary size of the model, $G$ is the group size for grouped query attention~\cite{ainslie2023gqa}, and in case of Llama3.1-8B, $G=\frac{d_k}{h_q}=\frac{128}{32}=4$. During inference, the activations for the current layer are processed while previously generated activations are discarded, rendering the total memory required to hold activations minuscule compared to other memory components. We exclude $ M_{\text{acts}} $ in our next calculation for brevity. For Llama3.1-8B which has 8 billion parameters with \texttt{float16} (2 bytes) precision for every component, the memory required to generate $N=96$ new tokens with a given context of size $32000$ tokens, will be
\begin{multline*}
\underbrace{8 \times 10^9}_{\text{Model}} + 
\underbrace{\left( 2 \times 32 \times \left( \frac{(96+32000) \times 4096}{4} \right) \right)}_{M_{\text{KV}}}\\+
\underbrace{\left(4096 \times 128256\right)}_{M_\text{logit}}
\end{multline*}
$\approx 21.22$ GiB. The $M_{\text{p}} $, $M_{KV}$, and, $M_{logit}$ are roughly 16.06, 4.11, and, 1.05 GiB, respectively. Due to the Grouped Query Attention, the size of $M_{KV}$ is reduced 4x times, making it feasible to work with very long context.

\subsection{Additional Detail on Metrics for QA Tasks}
Formally, let \( \text{Pred} = (p_1, p_2, \ldots, p_n) \) be the list of predicted tokens and \( \text{Ground} = (g_1, g_2, \ldots, g_m) \) be the list of ground-truth tokens. If \( C_P(x) \text{ and } C_G(x)\) are count of element $x$ in Pred and Ground respectively. The F1 score is defined on  \( s= \sum\limits_{x \in P \cap G}^{} \min(C_P(x), C_G(x))\) as  \(2 \times \frac{ \text{precision} \times \text{recall}}{\text{precision} + \text{recall}} \) where precision \( = \frac{s}{|\text{Pred}|} \) and recall \(= \frac{s}{|\text{Ground}|} \). The QA-F1 score is the F1 score computed after converting tokens to lowercase, removing punctuation, newlines, articles (a, an, the), and extra whitespace.

\section{Related Work}

Recent research on LLM optimization has largely confined its scope to isolated dimensions such as task accuracy or hardware efficiency, seldom addressing their intricate interplay. For instance, Li et al.~\cite{li2024evaluatingquantizedlargelanguage,NEURIPS2024_f89221ed} focus on quantization but primarily track accuracy metrics without system-level profiling. Lee et al.~\cite{lee2024comprehensiveevaluationquantizedinstructiontuned} extend these quantization evaluations to larger models yet maintain a similar emphasis on model accuracy. Jin et al.~\cite{jin-etal-2024-comprehensive} perform a systematic quantization analysis but restrict their study to a single optimization technique, whereas Yuan et al.~\cite{yuan-etal-2024-kv} explore efficient long-context handling for LLMs by examining model performance alone, overlooking hardware implications.

Beyond these specific studies, a broader landscape of LLM optimizations covers quantization~\cite{dettmers2022llmint8, frantar2022gptq, xiao2023awq}, sparse or low-rank techniques~\cite{hu2022lora,dettmers2023spqr}, and system-level methods~\cite{chen2023flexgen,dao2022flashattention}—often with a similar, narrow focus on either accuracy or throughput. For example, Dettmers et al.~\cite{dettmers2022llmint8} and Xiao et al.~\cite{xiao2023awq} analyze 8-bit quantization strategies but do not consider how quantization interacts with other optimizations like pruning. Other works, such as FlexGen~\cite{chen2023flexgen}, center on memory offloading and tiling strategies, providing valuable system-level insights yet neglecting trade-offs among different task types. Likewise, sparsity-centric solutions~\cite{hu2022lora,dettmers2023spqr} improve efficiency but typically examine only one form of compression at a time.

This fragmentation leaves several unanswered questions about how multiple techniques—e.g., pruning plus quantization—might jointly affect accuracy and efficiency across varied long-context tasks. For example, while pruning might boost summarization throughput, its effect on cross-document reasoning (e.g., MDQA) remains under-explored. Moreover, whether context length must be considered when pruning or quantizing is often overlooked, leading to potentially suboptimal deployment strategies.

In contrast, our work offers a multi-level optimization framework, jointly measuring memory, latency, and task-specific accuracy under different pruning, quantization, and token-level dropping strategies. Profiling these techniques up to 128K-token contexts reveals how individually beneficial methods can diminish or conflict when combined. We also demonstrate scalability on a 70B-parameter model, achieving consistent efficiency gains despite higher resource demands. Unlike prior studies, we provide a broader scope—covering multiple techniques and tasks—and deliver actionable insights for balancing performance, efficiency, and reliability in long-context LLM deployments.
\section{Extended Results and Discussion}
\subsection{Results of level-1}

\begin{table*}
\centering
\resizebox{\textwidth}{!}{%
\begin{tabular}{m{0.1cm}m{1.6cm}m{1.15cm}m{1.1cm}m{1.07cm}m{1.51cm}m{1.15cm}m{1.1cm}m{1.07cm}m{1.51cm}m{1.15cm}m{1.1cm}m{1.07cm}m{1.51cm}}
   &&\multicolumn{4}{c}{Qasper} & \multicolumn{4}{c}{HotpotQA} & \multicolumn{4}{c}{Gov Report} \\
   \cmidrule[\heavyrulewidth](lr){1-2} \cmidrule[\heavyrulewidth](lr){3-6} \cmidrule[\heavyrulewidth](lr){7-10} \cmidrule[\heavyrulewidth](lr){11-14}
   & \textbf{Method} & Score $\uparrow$ & Tok/s $\uparrow$  & Mem $\downarrow$ & No. Tok $\downarrow$  & Score $\uparrow$ & Tok/s $\uparrow$  & Mem $\downarrow$ & No. Tok $\downarrow$ & Score $\uparrow$ & Tok/s $\uparrow$  & Mem $\downarrow$ & No. Tok $\downarrow$ \\
   \cmidrule(lr){1-2} \cmidrule(lr){3-6} \cmidrule(lr){7-10} \cmidrule(lr){11-14} 
   \parbox[t]{0mm}{\multirow{7}{*}{\rotatebox[origin=c]{90}{\textbf{Llama3.1-8B}}}}
    & FP16 & 0.144 & 29.45 & \textbf{\textcolor{red}{17.75}} & 189.49 & 0.116 & 19.87 & \textbf{\textcolor{red}{18.61}} & 66.76 & \textbf{\textcolor{green}{0.306}} & 29.93 & \textbf{\textcolor{red}{18.27}} & \textbf{\textcolor{red}{490.41}}\\
    & BF16 & 0.149 & 29.41 & \textbf{\textcolor{red}{17.75}} & 191.49 & 0.118 & 20.05 & \textbf{\textcolor{red}{18.61}} & 61.26 & 0.305 & 29.17 & \textbf{\textcolor{red}{18.27}} & 488.05\\
    & FP16\_c & 0.115 & 25.26 & 16.91 & 209.83  & 0.117 & \textbf{\textcolor{red}{12.40}} & 17.31 & 56.50 & 0.286 & 28.86 & 17.12 & 484.47\\
    & BF16\_c & \textbf{\textcolor{red}{0.107}} & 25.65 & 16.91 & \textbf{\textcolor{red}{213.77}} & 0.119 & 13.38 & 17.31 & 62.08 & 0.283 & 28.23 & 17.12 & 484.67\\
    & KV-Q & 0.146 & 25.27 & 17.14 & 190.56 & 0.121 & 17.06 & 17.70 & 68.23 & 0.304 & 23.09 & 17.51 & 487.05\\
    & W4A16 & 0.140 & \textbf{\textcolor{green}{37.18}} & \textbf{\textcolor{green}{8.49}} & 188.27  & \textbf{\textcolor{red}{0.104}} & \textbf{\textcolor{green}{25.15}} & \textbf{\textcolor{green}{9.34}} & \textbf{\textcolor{red}{80.48}} & 0.302 & \textbf{\textcolor{green}{38.54}} & \textbf{\textcolor{green}{9.00}} & 473.34 \\
    & Minitron & \textbf{\textcolor{green}{0.238}} & \textbf{\textcolor{red}{24.88}} & 10.45 & \textbf{\textcolor{green}{56.70}} & \textbf{\textcolor{green}{0.331}} & 14.60 & 11.19 & \textbf{\textcolor{green}{54.10}} & \textbf{\textcolor{red}{0.065}} & \textbf{\textcolor{red}{19.10}} & 10.89 & \textbf{\textcolor{green}{164.43}} \\
    \cmidrule(lr){1-2} \cmidrule(lr){3-6} \cmidrule(lr){7-10} \cmidrule(lr){11-14} 

    \parbox[t]{0mm}{\multirow{7}{*}{\rotatebox[origin=c]{90}{\textbf{Mistral-Nemo-12B}}}}
    & FP16 & 0.129 & 21.71 & \textbf{\textcolor{red}{26.53}} & 164.22 & 0.084 & 15.19 & \textbf{\textcolor{red}{27.59}} & 85.96 & 0.244 & 21.99 & \textbf{\textcolor{red}{27.17}} & 338.39  \\
    & BF16 & \textbf{\textcolor{red}{0.127}} & 22.05 & 26.50 & 164.22 & 0.084 & 15.57 & 27.57 & 95.47 & 0.245 & 22.11 & 27.15 & 334.33\\
    & FP16\_c & 0.128 & 18.98 & 25.54 & \textbf{\textcolor{red}{166.54}} & 0.120 & 9.80 & 26.06 & 57.26 & \textbf{\textcolor{green}{0.250}} & 22.66 & 25.83 & 376.15  \\
    & BF16\_c & 0.128 & \textbf{\textcolor{red}{18.72}} & 25.53 & 159.65 & \textbf{\textcolor{green}{0.141}} & \textbf{\textcolor{red}{9.25}} & 26.04 & \textbf{\textcolor{green}{55.80}} & 0.247 & 22.80 & 25.82 & \textbf{\textcolor{red}{380.38}} \\
    & KV-Q & 0.139 & 18.74 & 24.52 & 157.71 & 0.079 & 13.26 & 24.54 & 90.89 & 0.246 & \textbf{\textcolor{red}{18.56}} & 24.54 & 338.59\\
    & W4A16 & 0.139 & \textbf{\textcolor{green}{27.35}} & \textbf{\textcolor{green}{11.45}} & \textbf{\textcolor{green}{156.33}} & \textbf{\textcolor{red}{0.074}} & \textbf{\textcolor{green}{18.51}} & \textbf{\textcolor{green}{12.53}} & 95.31 & 0.235 & \textbf{\textcolor{green}{27.78}} & \textbf{\textcolor{green}{12.10}} & 325.19 \\
    & Minitron & \textbf{\textcolor{green}{0.184}} & 23.22 & 18.84 & 159.13 & 0.137 & 18.13 & 19.89 & \textbf{\textcolor{red} {302.22}} & \textbf{\textcolor{red}{0.133}} & 22.83 & 19.47 & \textbf{\textcolor{green}{274.21}}\\
    \toprule
    
    & \textbf{Method} & \multicolumn{4}{c}{NarrativeQA} & \multicolumn{4}{c}{2WikiMQA} & \multicolumn{4}{c}{Multi News} \\
    \cmidrule(lr){1-2} \cmidrule(lr){3-6} \cmidrule(lr){7-10} \cmidrule(l){11-14}
    \parbox[t]{0mm}{\multirow{7}{*}{\rotatebox[origin=c]{90}{\textbf{Llama3.1-8B}}}}
    & FP16 & 0.096 & 11.13 & \textbf{\textcolor{red}{23.11}} & 160.26 & 0.292 & 13.22 & \textbf{\textcolor{red}{18.46}} & 16.43  & 0.241 & 31.46 & \textbf{\textcolor{red}{18.00}} & 416.83 \\
    & BF16 & 0.106 & 11.27 & 23.11 & 166.86 & 0.301 & 13.71 & \textbf{\textcolor{red}{18.46}} & 17.01 & \textbf{\textcolor{green}{0.242}} & 30.34 & 18.00 & 416.51\\
    & FP16\_c & \textbf{\textcolor{green}{0.115}} & 7.68 & 19.64 & 81.52 & \textbf{\textcolor{green}{0.322}} & 5.17 & 17.22 & \textbf{\textcolor{green}{13.29}} & 0.231 & 31.50 & 16.98 & 413.54\\
    & BF16\_c & 0.111 & \textbf{\textcolor{red}{6.82}} & 19.64 & \textbf{\textcolor{green}{81.31}} & 0.321 & \textbf{\textcolor{red}{4.65}} & 17.22 & 13.56 & 0.230 & 30.75 & 16.98 & 411.85\\
    & KV-Q & 0.099 & 9.32 & 20.63 & 164.13 & 0.270 & 12.75 & 17.59 & 21.00 & 0.240 & 25.48 & 17.33 & \textbf{\textcolor{red}{418.09}}\\
    & W4A16 & 0.100 & \textbf{\textcolor{green}{13.13}} & \textbf{\textcolor{green}{13.80}} & 159.78 & \textbf{\textcolor{red}{0.209}} & \textbf{\textcolor{green}{16.20}} & \textbf{\textcolor{green}{9.19}} & 20.41 & 0.238 & \textbf{\textcolor{green}{38.82}} & \textbf{\textcolor{green}{8.74}} & 390.74\\
    & Minitron & \textbf{\textcolor{red}{0.084}} & 11.43 & 14.95 & \textbf{\textcolor{red}{273.24}} & 0.305 & 15.15 & 11.06 & \textbf{\textcolor{red}{41.92}} & \textbf{\textcolor{red}{0.092}} & \textbf{\textcolor{red}{24.91}} & 10.67 & \textbf{\textcolor{green}{201.06}} \\
    \cmidrule(lr){1-2} \cmidrule{3-14}

        \parbox[t]{0mm}{\multirow{7}{*}{\rotatebox[origin=c]{90}{\textbf{Mistral-Nemo-12B}}}}
    & FP16 & 0.025 & 8.46 & 33.36 & 192.90 & 0.144 & 10.67 & \textbf{\textcolor{red}{27.39}} & 50.37 & 0.222 & 22.32 & \textbf{\textcolor{red}{26.98}} & 293.38\\
    & BF16 & 0.023 & 9.11 & \textbf{\textcolor{red}{33.37}} & 229.12 & 0.147 & 10.90 & 27.37 & 51.21 & 0.220 & 22.23 & 26.96 & 283.31 \\
    & FP16\_c & 0.057 & 7.05 & 28.83 & \textbf{\textcolor{green}{135.00}} & \textbf{\textcolor{green}{0.250}} & 4.23 & 25.96 & 15.40 & 0.220 & 22.87 & 25.65 & \textbf{\textcolor{red}{296.44}} \\
    & BF16\_c & \textbf{\textcolor{green}{0.062}} & \textbf{\textcolor{red}{6.89}} & 28.84 & 138.57 & 0.209 & \textbf{\textcolor{red}{3.93}} & 25.93 & \textbf{\textcolor{green}{14.41}} & 0.216 & 22.63 & 25.63 & 291.70\\
    & KV-Q & 0.024 & 7.08 & 24.52 & 207.78 & 0.140 & 9.77 & 24.52 & 47.67 & \textbf{\textcolor{green}{0.223}} & \textbf{\textcolor{red}{18.66}} & 24.52 & 280.09\\
    & W4A16 & 0.023 & 10.39 & \textbf{\textcolor{green}{18.30}} & 210.96  & \textbf{\textcolor{red}{0.128}} & 13.43 & \textbf{\textcolor{green}{12.32}} & 53.24 & 0.221 & \textbf{\textcolor{green}{27.98}} & \textbf{\textcolor{green}{11.91}} & 267.74 \\
    & Minitron & \textbf{\textcolor{red}{0.018}} & \textbf{\textcolor{green}{12.96}} & 25.35 & \textbf{\textcolor{red}{491.18}} & 0.168 & \textbf{\textcolor{green}{17.52}} & 19.71 & \textbf{\textcolor{red}{234.15}} & \textbf{\textcolor{red}{0.118}} & 23.03 & 19.31 & \textbf{\textcolor{green}{233.80}}\\
    \bottomrule
    
\end{tabular} }
    \caption{Evaluation Results for Level 1 Optimization}
     \label{tab:all_lvl1_metrics}
\end{table*}

\begin{figure*}[h]
\centering
\caption{Comparative Performance of Level 1 Optimization Methods vs. FP16 Baseline for \llamaModel8B and Mistral Nemo}
\begin{tabular}{cccc}
\includegraphics[width=0.23\textwidth]{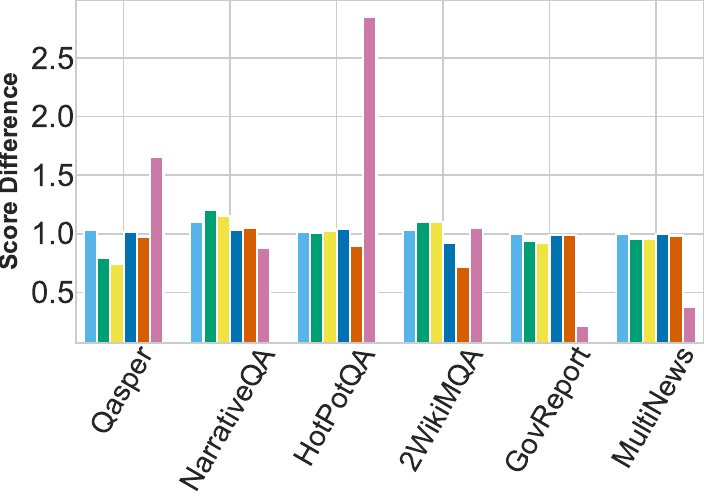} &
\includegraphics[width=0.23\textwidth]{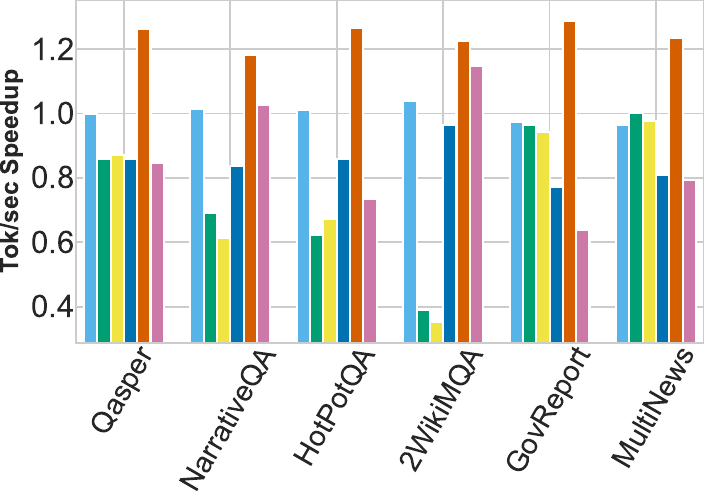} &
\includegraphics[width=0.23\textwidth]{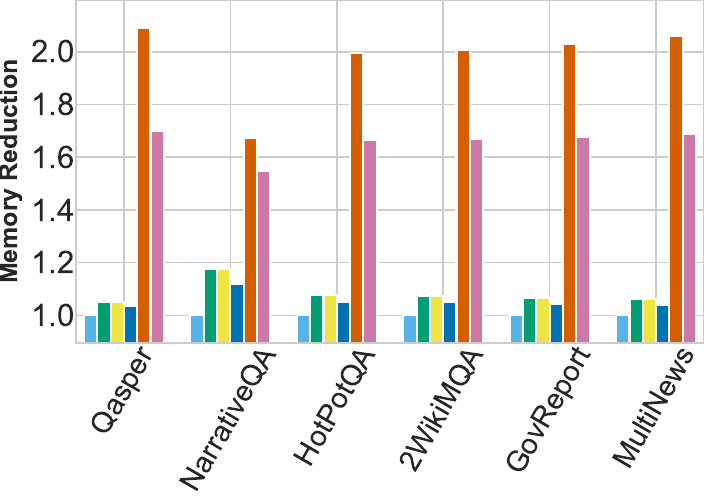} &
\includegraphics[width=0.23\textwidth]{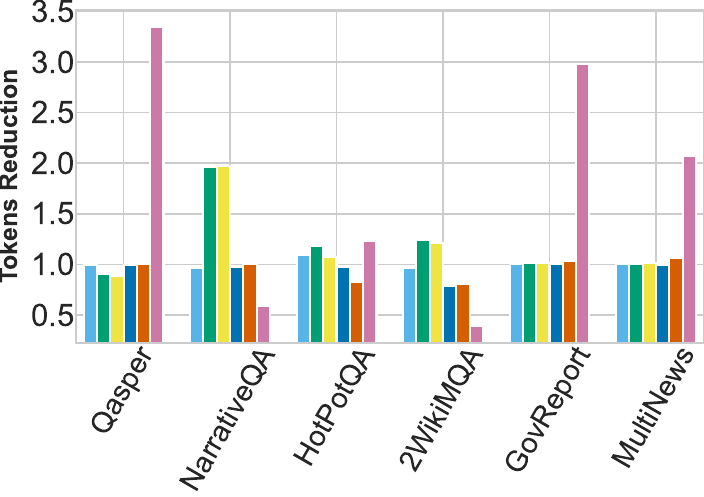} \\
\includegraphics[width=0.23\textwidth]{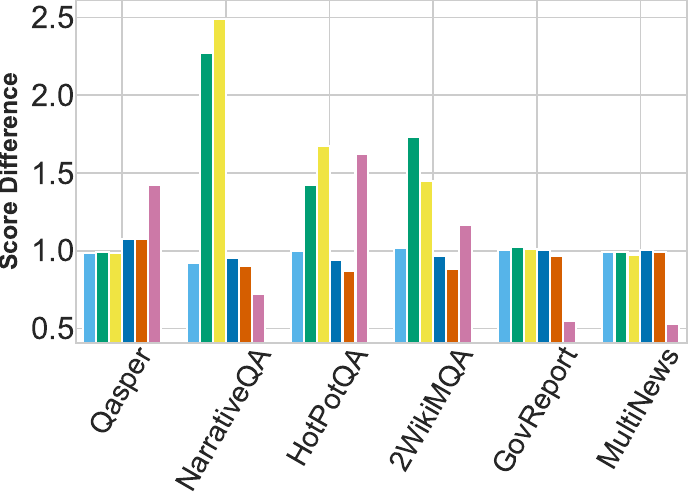} &
\includegraphics[width=0.23\textwidth]{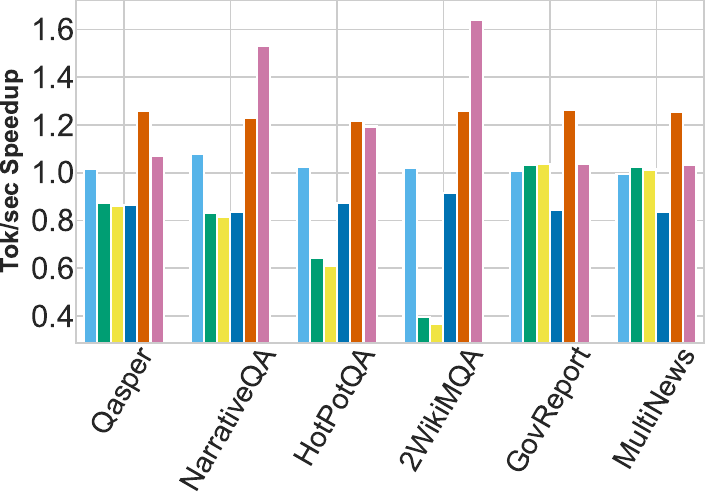} &
\includegraphics[width=0.23\textwidth]{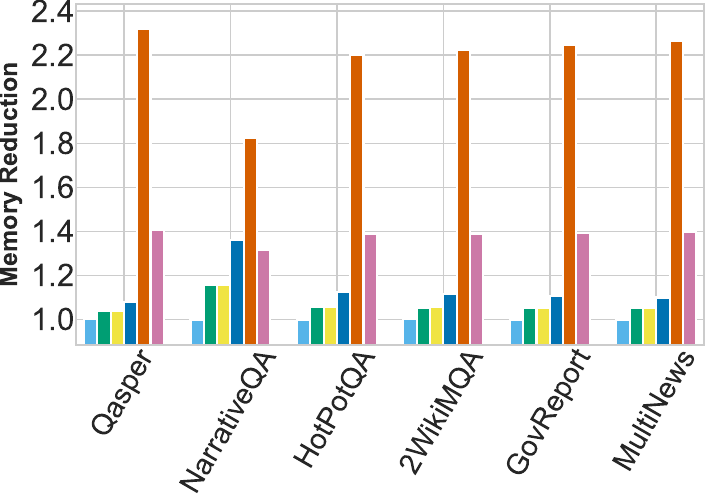} &
\includegraphics[width=0.23\textwidth]{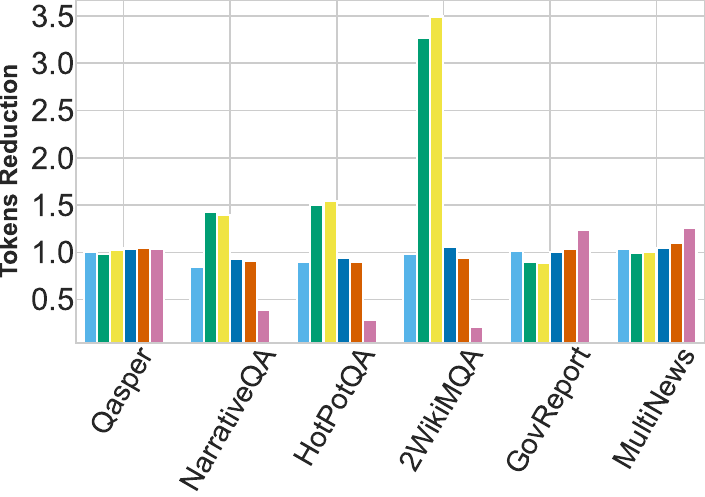} \\
\multicolumn{4}{c}{
  \includegraphics[width=0.7\textwidth]{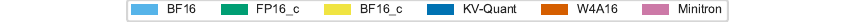}
} \\
\end{tabular}
\label{fig:all_perf_diff_lvl1}
\end{figure*}

Table~\ref{tab:all_lvl1_metrics} summarizes system‑level performance and text‑quality results for all tasks using the level‑1 inference‑optimization methods on both \llamaModel8B and Mistral‑NeMo. Table~\ref{tab:all_lvl1_f1} reports the corresponding text‑quality metrics for the SDQA and MDQA long‑context tasks, while Figure~\ref{fig:all_perf_diff_lvl1} depicts the performance gains achieved by each Level‑1 optimization relative to the baseline.

\begin{figure*}[t]
    \caption{Performance Across Level-2 Optimization Methods for \llamaModel 8B and Mistral-Nemo}
  \includegraphics[width=0.245\textwidth,height=4.1cm]{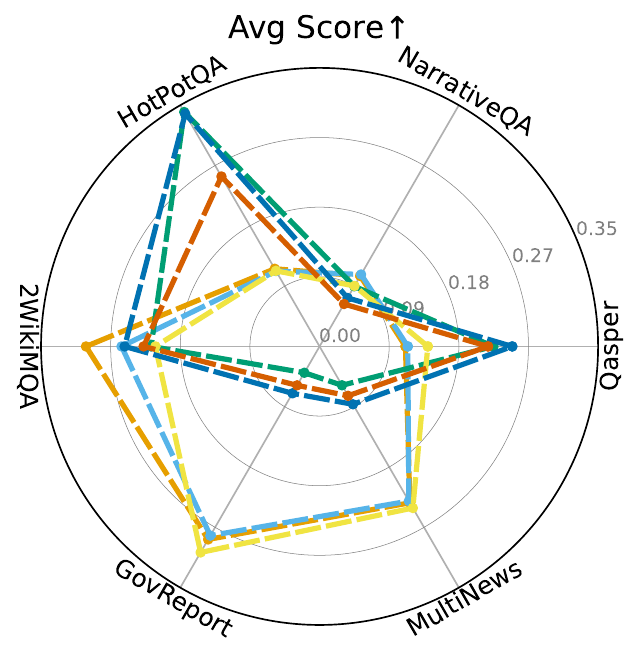}
  \includegraphics[width=0.245\textwidth,height=4.1cm]{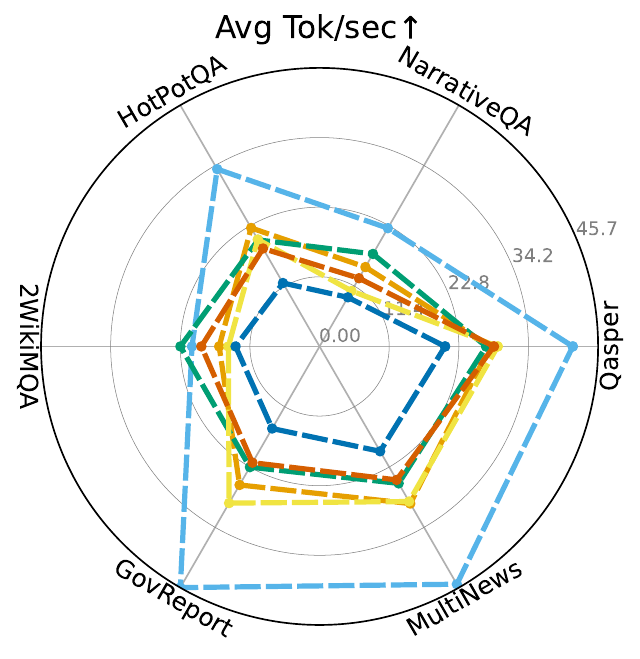}
  \includegraphics[width=0.245\textwidth,height=4.1cm]{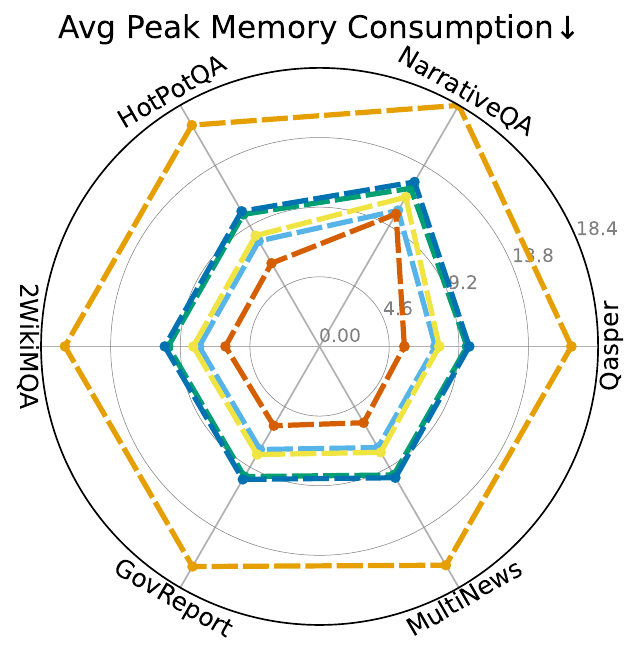}
  \includegraphics[width=0.245\textwidth,height=4.1cm]{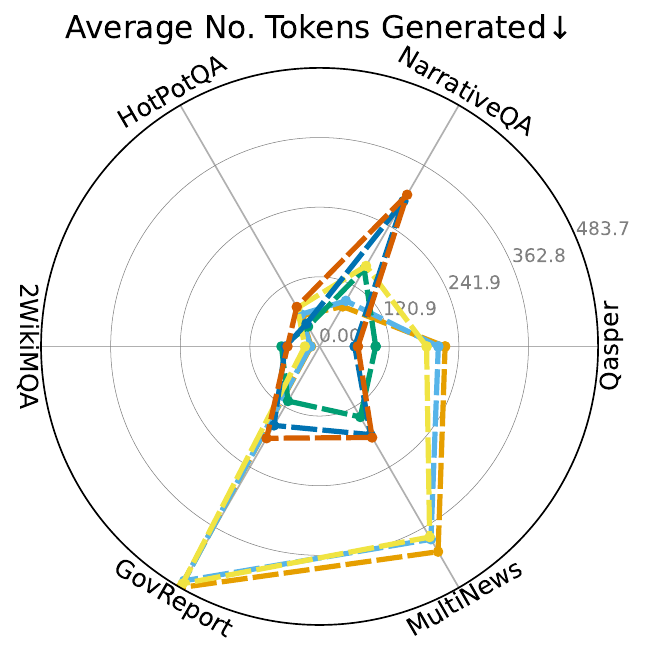}
  \includegraphics[width=0.245\textwidth,height=4.1cm]{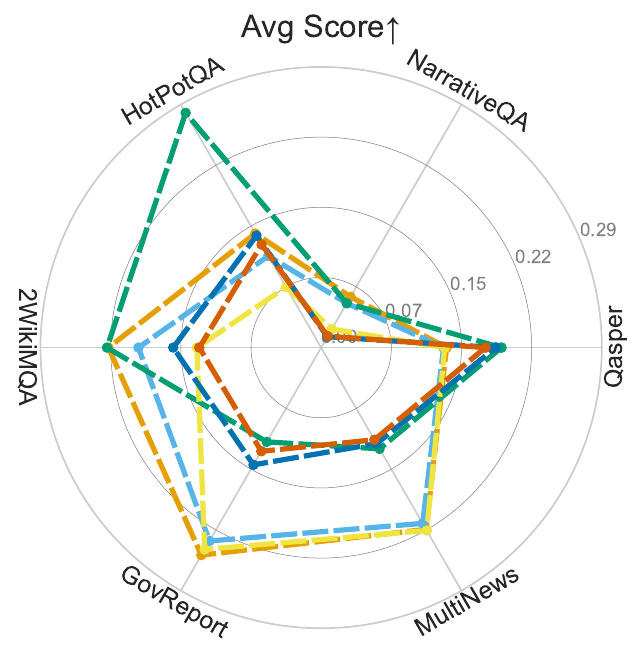}
  \includegraphics[width=0.245\textwidth,height=4.1cm]{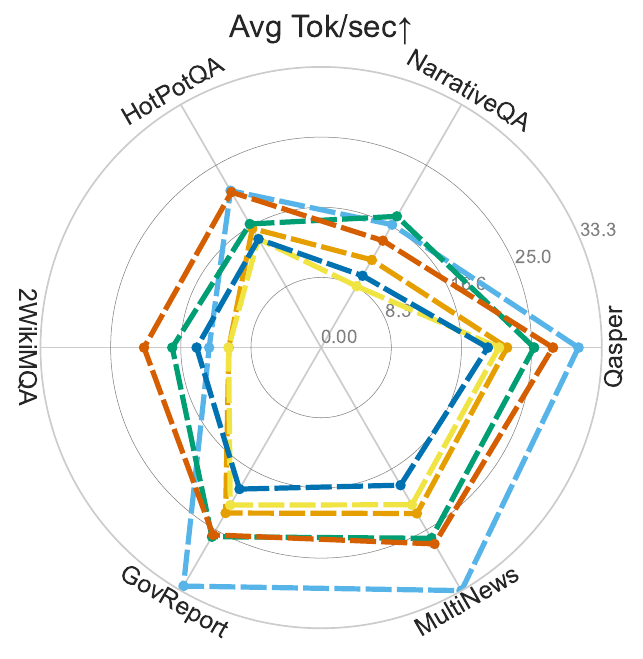}
  \includegraphics[width=0.245\textwidth,height=4.1cm]{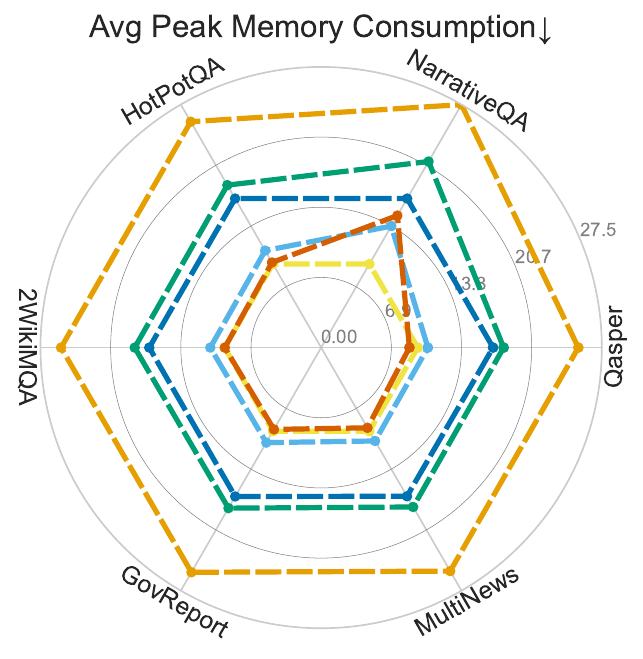}
  \includegraphics[width=0.245\textwidth,height=4.1cm]{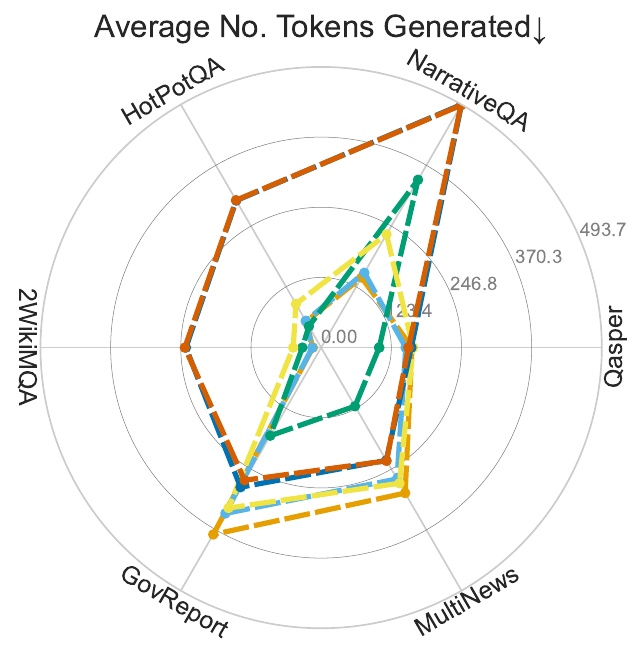}
  \centering
  \includegraphics[width=0.8\textwidth,height=0.6cm]{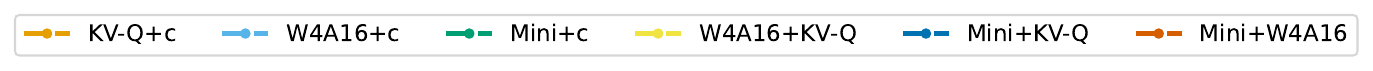}
  \label{fig:radar_all_lvl2_all}
\end{figure*}

\begin{figure*}[t]
\centering
\caption{Comparative Scalability Performance of Level 1 Optimization Methods vs. FP16 Baseline}
\begin{tabular}{@{\hspace*{-0.4em}}cccc}
\includegraphics[width=0.23\textwidth]{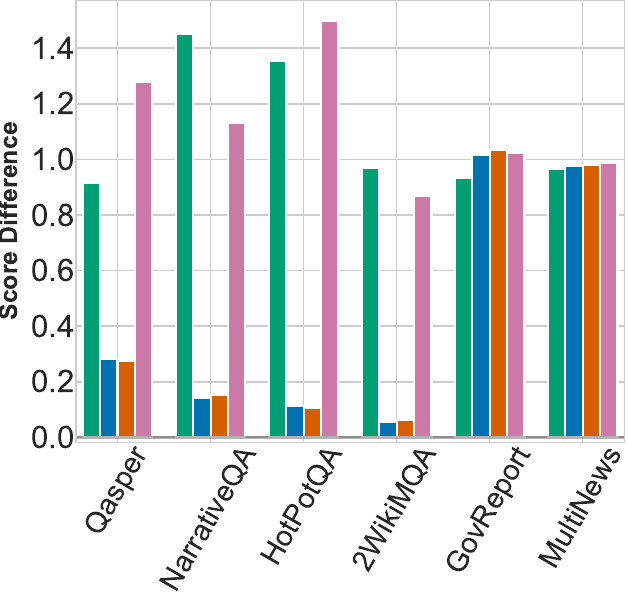} &
\includegraphics[width=0.23\textwidth]{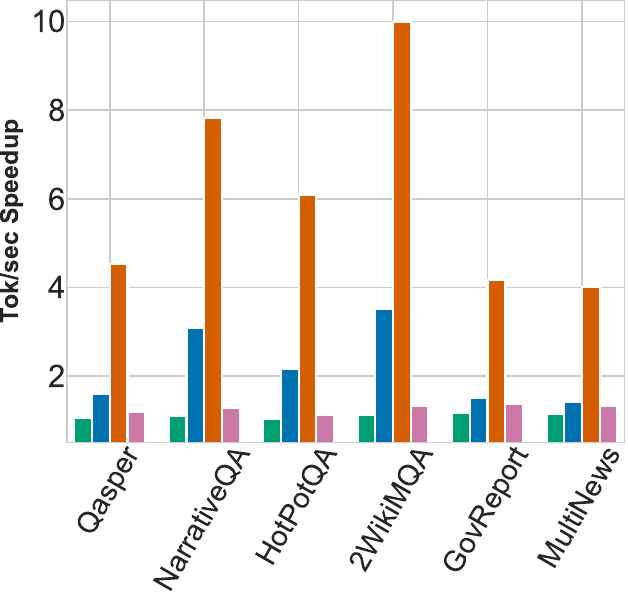} &
\includegraphics[width=0.23\textwidth]{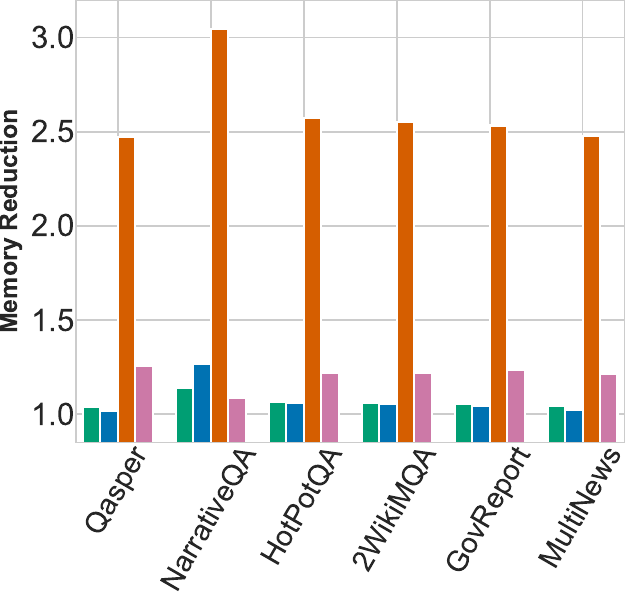} &
\includegraphics[width=0.23\textwidth]{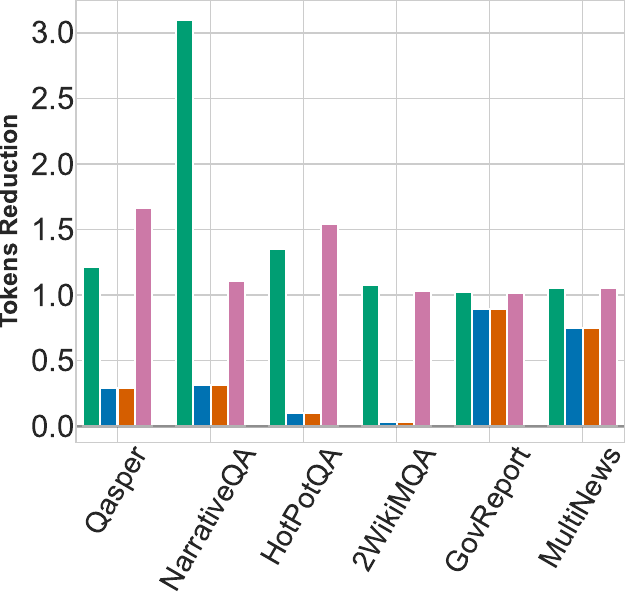}   \\
\multicolumn{4}{c}{
  \includegraphics[width=0.7\textwidth]{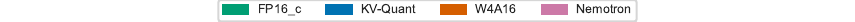}
} \\
\end{tabular}
\label{fig:all_perf_diff_lvl70b}
\end{figure*}

Figure~\ref{fig:radar_all_lvl2_all} summarizes system‑level performance and text quality results for all tasks using the Level‑2 inference‑optimization methods on both \llamaModel8B and Mistral‑NeMo as radar plot and shows the raw values as in Table~\ref{tab:all_lvl2_metrics}. Table~\ref{tab:all_lvl2_f1} reports the corresponding text‑quality metrics for the SDQA and MDQA long‑context tasks, while Figure~\ref{fig:all_perf_diff_lvl70b} depicts the performance gains relative to the baseline for model on a scale with 70 billion parameters when inference optimization are applied.

\textbf{Observation 7 underscores the task‑specific nature of inference‑time optimizations.}
For \llamaModel8B, the combination of weight pruning with token dropping (Mini+ C) delivers the highest overall accuracy, increasing the mean score across six benchmarks by 10.23\%. This gain is driven by a 50\% improvement on four question‑answering (QA) tasks, although it is offset by a 71.13\% decline on the two summarization tasks. By contrast, Mistral‑NeMo attains its best results with weight pruning plus key–value quantization (Mini+KV‑Q), achieving a 60.28\% average boost including a 115.14\% jump in QA performance  but still suffering a 49.44\% drop in summarization. These findings confirm that pruned networks are poorly suited to summarization and illustrate why selecting an optimal inference‑optimization pipeline is both nuanced and highly task‑dependent. Practical deployments therefore require an intelligent scheduler that first identifies the task type and then applies the optimization strategy most likely to maximize end‑to‑end performance.

\textbf{Performing prompt compression on GPU is unfavorable } While the prompt compression is faster on the GPU as in our experiments it takes 0.68 seconds to compress a 5400 word prompt versus 2.56 seconds on the CPU, it costs an extra $\approx$ 1.14 GB of VRAM to host the model the drop the tokens and the tokens themselves. For long-context workloads that are already memory-bound, this trade-off is unfavorable, so we recommend running prompt compression on the CPU.

\begin{table*}
\centering
\resizebox{\textwidth}{!}{%
\begin{NiceTabular}{|m{0.1cm}|m{2.0cm}|>{\centering\arraybackslash}m{1.5cm}|>{\centering\arraybackslash}m{1.3cm}|>{\centering\arraybackslash}m{1.4cm}|>{\centering\arraybackslash}m{2.0cm}|>{\centering\arraybackslash}m{1.5cm}|>{\centering\arraybackslash}m{1.3cm}|>{\centering\arraybackslash}m{1.4cm}|>{\centering\arraybackslash}m{2.0cm}|}
\toprule
\Block{1-*}{}
   & &\multicolumn{4}{c}{\textbf{Qasper}} & \multicolumn{4}{c}{\textbf{HotpotQA}} \\
   \cmidrule[\heavyrulewidth](lr){1-2} \cmidrule[\heavyrulewidth](lr){3-6} \cmidrule[\heavyrulewidth](lr){7-10}
    & \textbf{Method} & F1 QA $\uparrow$ & Prec $\uparrow$ & Recall $\uparrow$ & Hall. Score $\downarrow$ & F1 QA $\uparrow$ & Prec $\uparrow$ & Recall $\uparrow$ & Hall. Score $\downarrow$ \\
   \cmidrule(lr){1-2} \cmidrule(lr){3-6} \cmidrule(lr){7-10}
   \parbox[t]{0mm}{\multirow{7}{*}{\rotatebox[origin=c]{90}{Llama3.1-8B}}}    
    & FP16 & 14.41 & 9.22 & 62.67 & 18.98 & 11.59 & 6.84 & 66.04 & 11.52 \\
    & BF16 & 14.92 & 9.43 & 63.51 & 19.23 & 11.76 & 6.96 & 64.79 & 11.40\\
    & FP16\_c & 11.48 & 6.90 & 58.96 & 17.15 & 11.73 & 6.99 & 60.13 & \textbf{\cellcolor{blue!15}{8.62}}\\
    & BF16\_c & \cellcolor{red!40}{10.70} & \cellcolor{red!40}{6.43} & 59.59 & 17.12 & 11.91 & 7.01 & 65.20 & 9.10\\
    & KV-Q & 14.58 & 9.24 & \textbf{\cellcolor{blue!15}{65.70}} & \cellcolor{red!40}{19.68} & 12.06 & 7.17 & \textbf{\cellcolor{blue!15}{66.43}} & 11.87\\
    & W4A16 & 13.97 & 8.69 & 62.53 & 17.76 & \cellcolor{red!40}{10.39} & \cellcolor{red!40}{6.02} & 65.88 & 12.73\\
    & Minitron & \textbf{\cellcolor{blue!15}{23.82}} & \textbf{\cellcolor{blue!15}{21.82}} & \cellcolor{red!40}{38.50} & \textbf{\cellcolor{blue!15}{16.60}} & \textbf{\cellcolor{blue!15}{33.08}} & \textbf{\cellcolor{blue!15}{33.50}} & \cellcolor{red!40}{44.27} & \cellcolor{red!40}{16.62}\\
    \cmidrule(lr){1-2} \cmidrule(lr){3-6} \cmidrule(lr){7-10}

    \parbox[t]{0mm}{\multirow{7}{*}{\rotatebox[origin=c]{90}{Mistral-Nemo-12B}}}
    & FP16 & 12.92 & 8.05 & 54.55 & 22.21 & 8.45 & 5.19 & 53.69 & 16.86 \\
    & BF16 & \cellcolor{red!40}{12.72} & 7.95 & 54.82 & 22.12 & 8.42 & 5.03 & \textbf{\cellcolor{blue!15}{57.43}} & 17.07 \\
    & FP16\_c & 12.83 & \cellcolor{red!40}{7.91} & 55.30 & 18.91 & 12.01 & 7.97 & 56.76 & 12.17\\
    & BF16\_c & 12.77 & 7.97 & 54.09 & \textbf{\cellcolor{blue!15}{18.62}} & \textbf{\cellcolor{blue!15}{14.13}} & 10.04 & 57.20 & \textbf{\cellcolor{blue!15}{11.22}} \\
    & KV-Q & 13.87 & 8.69 & 56.32 & 22.59  & 7.93 & 5.00 & 51.73 & 16.21 \\
    & W4A16 & 13.94 & 8.58 & \textbf{\cellcolor{blue!15}{58.20}} & 20.06 & \cellcolor{red!40}{7.38} & \cellcolor{red!40}{4.29} & 47.47 & 16.88 \\
    & Minitron & \textbf{\cellcolor{blue!15}{18.40}} & \textbf{\cellcolor{blue!15}{13.80}} & \cellcolor{red!40}{51.48} & \cellcolor{red!40}{28.93} & 13.71 & \textbf{\cellcolor{blue!15}{13.24}} & \cellcolor{red!40}{32.33} & \cellcolor{red!40}{49.15} \\
    \bottomrule 
    & \textbf{Method} & \multicolumn{4}{c}{\textbf{NarrativeQA}} & \multicolumn{4}{c}{\textbf{2WikiMQA}} \\
    \cmidrule(lr){1-2} \cmidrule(lr){3-6} \cmidrule(lr){7-10}
    \parbox[t]{0mm}{\multirow{7}{*}{\rotatebox[origin=c]{90}{Llama3.1-8B}}}
    & FP16 & 9.57 & \cellcolor{red!40}{6.20} & 44.96 & 22.75 & 29.18 & 25.83 & 42.33 & 92.46\\
    & BF16 & 10.57 & 6.78 & 45.80 & 23.07 & 30.14 & 26.82 & 44.62 & 91.96\\
    & FP16\_c & \textbf{\cellcolor{blue!15}{11.49}} & 7.68 & 45.27 & \textbf{\cellcolor{blue!15}{17.60}} & \textbf{\cellcolor{blue!15}{32.19}} & \textbf{\cellcolor{blue!15}{30.96}} & \cellcolor{red!40}{37.42} & 88.28 \\
    & BF16\_c & 11.06 & 7.27 & 43.99 & 18.20 & 32.11 & 30.81 & 37.91 & 88.26 \\
    & KV-Q & 9.92 & 6.48 & 45.32 & \cellcolor{red!40}{24.86} & 27.03 & 22.93 & \textbf{\cellcolor{blue!15}{46.22}} & \cellcolor{red!40}{93.18}\\
    & W4A16 & 10.04 & 6.42 & \textbf{\cellcolor{blue!15}{48.83}} & 22.96 & \cellcolor{red!40}{20.91} & \cellcolor{red!40}{17.66} & 37.48 & 92.25\\
    & Minitron & \cellcolor{red!40}{8.38} & \textbf{\cellcolor{blue!15}{7.74}} & \cellcolor{red!40}{20.66} & 22.77  & 30.52 & 29.10 & 40.87 & \textbf{\cellcolor{blue!15}{20.04}} \\
      \cmidrule(lr){1-2} \cmidrule(lr){3-6} \cmidrule(lr){7-10}
    \parbox[t]{0mm}{\multirow{7}{*}{\rotatebox[origin=c]{90}{Mistral-Nemo-12B}}}
 & FP16 & 2.51 & 1.47 & 19.93 & 36.52 & 14.44 & 11.89 & 32.95 & 96.84\\
& BF16 & 2.32 & 1.43 & 22.42 & 38.96  & 14.72 & 12.50 & 34.11 & 96.86\\
& FP16\_c & 5.70 & 3.52 & 35.68 & 24.77  & \textbf{\cellcolor{blue!15}{24.98}} & \textbf{\cellcolor{blue!15}{22.44}} & \textbf{\cellcolor{blue!15}{41.84}} & 96.25\\
& BF16\_c & \textbf{\cellcolor{blue!15}{6.24}} & \textbf{\cellcolor{blue!15}{3.91}} & \textbf{\cellcolor{blue!15}{38.67}} & \textbf{\cellcolor{blue!15}{22.21}} & 20.94 & 19.69 & \cellcolor{red!40}{31.07} & 93.50\\
& KV-Q & 2.39 & 1.54 & 22.50 & 37.88 & 14.00 & 11.55 & 33.24 & 96.48 \\
& W4A16 & 2.27 & 1.50 & 20.99 & 37.52 & \cellcolor{red!40}{12.79} & \cellcolor{red!40}{10.27} & 32.75 & \cellcolor{red!40}{97.19} \\
& Minitron & \cellcolor{red!40}{1.82} & \cellcolor{red!40}{1.39} & \cellcolor{red!40}{11.11} & \cellcolor{red!40}{67.54} & 16.84 & 14.43 & 39.21 & \textbf{\cellcolor{blue!15}{43.98}}\\
\bottomrule
\end{NiceTabular} }
    \caption{Accuracy comparison for level 1. Left column is SDA and right column is MDA.}
     \label{tab:all_lvl1_f1}
\end{table*}

\textbf{Textual prompt compression delivers minimal efficiency improvements for long-context processing while introducing significant performance tradeoffs across most tasks}. Despite its theoretical appeal, this approach achieves only modest memory savings (1.08x reduction in memory consumption) compared to more effective quantization and pruning methods, while simultaneously degrading throughput to 0.7x of baseline performance. Our experimental implementation, which executes prompt compression operations on the CPU, frequently results in longer text generation times that exceed the baseline model. While performance metrics generally show no significant improvements, MDQA scenarios represent a potential application where prompt compression could provide benefits, particularly when synthesizing information from multiple research papers or technical documents. However, even in these specialized cases, the marginal advantages should be carefully weighed against increased computational overhead and limited memory optimization benefits. Comprehensive metric comparisons across all tested approaches are available in Table~\ref{tab:all_lvl1_metrics}.

\textbf{KV quantization's limited memory savings come at the cost of lower throughput and a higher risk of hallucination, though it does provide strong recall capabilities for QA tasks.} KV quantization yields only a 1.06x reduction in memory consumption while decreasing throughput to 0.85x compared to the baseline, indicating that it may not offer sufficient performance benefits for LLMs using grouped query attention. In addition, it exhibits the highest hallucination score, suggesting an elevated likelihood of generating fabricated or incorrect content. Notably, however, KV quantization achieves the highest recall score across all QA tasks as shown in Table~\ref{tab:all_lvl1_f1} it effectively retrieves and presents relevant information from the dataset even if it occasionally produces content that is factually inaccurate.

\subsection{Results of level-2}

\begin{table*}
\centering
\resizebox{\textwidth}{!}{%
\begin{tabular}{m{0.1cm}m{2.0cm}m{1.15cm}m{1.1cm}m{1.07cm}m{1.51cm}m{1.15cm}m{1.1cm}m{1.07cm}m{1.51cm}m{1.15cm}m{1.1cm}m{1.07cm}m{1.51cm}}
   &&\multicolumn{4}{c}{Qasper} & \multicolumn{4}{c}{HotpotQA} & \multicolumn{4}{c}{Gov Report} \\
   \cmidrule[\heavyrulewidth](lr){1-2} \cmidrule[\heavyrulewidth](lr){3-6} \cmidrule[\heavyrulewidth](lr){7-10} \cmidrule[\heavyrulewidth](lr){11-14}
   & \textbf{Method} & Score $\uparrow$ & Tok/s $\uparrow$  & Mem $\downarrow$ & No. Tok $\downarrow$  & Score $\uparrow$ & Tok/s $\uparrow$  & Mem $\downarrow$ & No. Tok $\downarrow$ & Score $\uparrow$ & Tok/s $\uparrow$  & Mem $\downarrow$ & No. Tok $\downarrow$ \\
   \cmidrule(lr){1-2} \cmidrule(lr){3-6} \cmidrule(lr){7-10} \cmidrule(lr){11-14} 
   \parbox[t]{0mm}{\multirow{7}{*}{\rotatebox[origin=c]{90}{\textbf{Llama3.1-8B}}}}
    & FP16 & 14.41 & 9.22 & 62.67 & 18.98 & 11.59 & 6.84 & 66.04 & 11.52 & 0.306 & 29.93 & 18.27 & 490.41\\
    & KV-Q+c & \textbf{\textcolor{red}{0.109}} & 27.99 & \textbf{\textcolor{red}{16.63}} & \textbf{\textcolor{red}{218.02}} & 0.114 & 22.42 & \textbf{\textcolor{red}{16.87}} & 61.44 & 0.283 & 26.22 & \textbf{\textcolor{red}{16.78}} & \textbf{\textcolor{red}{483.70}} \\
    & W4A16+c & 0.112 & \textbf{\textcolor{green}{41.51}} & 7.65 & 206.24 & \textbf{\textcolor{red}{0.111}} & \textbf{\textcolor{green}{33.56}} & 8.03 & 64.37 & 0.278 & \textbf{\textcolor{green}{45.64}} & 7.86 & 469.93 \\
    & Mini+c & 0.221 & 27.31 & 9.74 & 97.47 & \textbf{\textcolor{green}{0.344}} & 20.16 & 10.07 & \textbf{\textcolor{green}{39.76}}  & \textbf{\textcolor{red}{0.039}} & 22.84 & 9.92 & \textbf{\textcolor{green}{109.53}}  \\
    & W4A16+KV\textsuperscript{Q} & 0.138 & 29.19 & 7.91 & 185.83 & 0.111 & 20.24 & 8.45 & 77.09 & \textbf{\textcolor{green}{0.302}} & 29.65 & 8.24 & 475.89  \\
    & Mini+KV\textsuperscript{Q} & \textbf{\textcolor{green}{0.245}} & \textbf{\textcolor{red}{20.57}} & 9.88 & \textbf{\textcolor{green}{61.05}} & 0.342 & \textbf{\textcolor{red}{12.00}} & 10.30 & 45.57 & 0.069 & \textbf{\textcolor{red}{15.54}} & 10.14 & 158.36\\
    & Mini+W4A16 & 0.214 & 28.57 & \textbf{\textcolor{green}{5.60}} & 66.55  & 0.249 & 18.54 & \textbf{\textcolor{green}{6.34}} & \textbf{\textcolor{red}{79.21}} & 0.057 & 22.00 & \textbf{\textcolor{green}{6.05}} & 184.10  \\
    \cmidrule(lr){1-2} \cmidrule(lr){3-6} \cmidrule(lr){7-10} \cmidrule(lr){11-14} 

    \parbox[t]{0mm}{\multirow{7}{*}{\rotatebox[origin=c]{90}{\textbf{Mistral-Nemo-12B}}}}
    & FP16 & 0.144 & 29.45 & 17.75 & 189.49 & 0.116 & 19.87 & 18.61 & 66.76 & 0.306 & 29.93 & 18.27 & 490.41\\
    & KV-Q+c & 0.130 & 22.05 & \textbf{\textcolor{red}{25.23}} & 159.77  & 0.136 & 16.32 & \textbf{\textcolor{red}{25.59}} & 52.89 & \textbf{\textcolor{green}{0.248}} & 22.66 & \textbf{\textcolor{red}{25.46}} & \textbf{\textcolor{red}{379.60}}\\
    & W4A16+c & \textbf{\textcolor{red}{0.127}} & \textbf{\textcolor{green}{30.52}} & 10.47 & 148.88 & 0.109 & \textbf{\textcolor{green}{21.49}} & 10.99 & 54.34 & 0.231 & \textbf{\textcolor{green}{32.66}} & 10.77 & 336.90 \\
    & Mini+c & \textbf{\textcolor{green}{0.186}} & 25.26 & 17.91 & \textbf{\textcolor{green}{102.05}} & \textbf{\textcolor{green}{0.280}} & 16.93 & 18.41 & \textbf{\textcolor{green}{43.20}} & \textbf{\textcolor{red}{0.113}} & 25.91 & 18.19 & \textbf{\textcolor{green}{178.91}}\\
    & W4A16+KV\textsuperscript{Q} & 0.128 & 21.09 & 9.47 & \textbf{\textcolor{red}{161.75}} & \textbf{\textcolor{red}{0.072}} & 14.91 & \textbf{\textcolor{green}{9.47}} & 88.67 & 0.241 & 21.56 & 9.47 & 325.82\\
    & Mini+KV\textsuperscript{Q} & 0.181 & \textbf{\textcolor{red}{19.76}} & 16.87 & 158.59 & 0.134 & \textbf{\textcolor{red}{14.87}} & 16.89 & 298.67 & 0.140 & \textbf{\textcolor{red}{19.38}} & 16.87 & 283.54   \\
    & Mini+W4A16 & 0.170 & 27.49 & \textbf{\textcolor{green}{8.66}} & 153.61 & 0.123 & 21.29 & 9.65 & \textbf{\textcolor{red}{299.30}} & 0.124 & 25.65 & \textbf{\textcolor{green}{9.27}} & 269.44 \\
    \toprule
    
    & \textbf{Method} & \multicolumn{4}{c}{NarrativeQA} & \multicolumn{4}{c}{2WikiMQA} & \multicolumn{4}{c}{Multi News} \\
    \cmidrule(lr){1-2} \cmidrule(lr){3-6} \cmidrule(lr){7-10} \cmidrule(l){11-14}
    \parbox[t]{0mm}{\multirow{7}{*}{\rotatebox[origin=c]{90}{\textbf{Llama3.1-8B}}}}
    & FP16 & 0.096 & 11.13 & 23.11 & 160.26 & 0.292 & 13.22 & 18.46 & 16.43 & 0.241 & 31.46 & 18.00 & 416.83 \\
    & KV-Q+c & 0.103 & 15.02 & \textbf{\textcolor{red}{18.38}} & \textbf{\textcolor{green}{80.14}} & \textbf{\textcolor{green}{0.297}} & 16.46 & \textbf{\textcolor{red}{16.82}} & \textbf{\textcolor{green}{14.77}} & 0.230 & 29.74 & \textbf{\textcolor{red}{16.68}} & \textbf{\textcolor{red}{411.59}}\\
    & W4A16+c & \textbf{\textcolor{green}{0.106}} & \textbf{\textcolor{green}{22.38}} & 10.37 & 91.36 & 0.252 & 20.91 & 7.95 & 15.85 & 0.227 & \textbf{\textcolor{green}{45.00}} & 7.71 & 386.76 \\
    & Mini+c & 0.088 & 17.51 & 12.06 & 154.90 & 0.214 & \textbf{\textcolor{green}{22.81}} & 9.99 & \textbf{\textcolor{red}{66.22}} & \textbf{\textcolor{red}{0.057}} & 25.92 & 9.80 & \textbf{\textcolor{green}{141.52}}\\
    & W4A16+KV\textsuperscript{Q} & 0.089 & 10.49 & 11.38 & 161.45 & \textbf{\textcolor{red}{0.209}} & 14.95 & 8.34 & 25.48 & \textbf{\textcolor{green}{0.237}} & 29.35 & 8.08 & 382.26\\
    & Mini+KV\textsuperscript{Q} & 0.071 & \textbf{\textcolor{red}{9.29}} & 12.54 & 291.19 & 0.248 & \textbf{\textcolor{red}{13.77}} & 10.22 & 59.30 & 0.085 & \textbf{\textcolor{red}{19.85}} & 10.01 & 176.86\\
    & Mini+W4A16 & \textbf{\textcolor{red}{0.062}} & 12.91 & \textbf{\textcolor{green}{10.10}} & \textbf{\textcolor{red}{304.18}} & 0.224 & 19.37 & \textbf{\textcolor{green}{6.22}} & 55.68 & 0.072 & 25.26 & \textbf{\textcolor{green}{5.82}} & 182.62 \\
    \cmidrule(lr){1-2} \cmidrule{3-14}

        \parbox[t]{0mm}{\multirow{7}{*}{\rotatebox[origin=c]{90}{\textbf{Mistral-Nemo-12B}}}}
    & FP16 & 0.025 & 8.46 & 33.36 & 192.90 & 0.144 & 10.67 & 27.39 & 50.37 & 0.222 & 22.32 & 26.98 & 293.38\\
    & KV-Q+c & \textbf{\textcolor{green}{0.061}} & 12.01 & \textbf{\textcolor{red}{27.53}} & \textbf{\textcolor{green}{142.35}} & 0.220 & \textbf{\textcolor{red}{10.95}} & \textbf{\textcolor{red}{25.52}} & \textbf{\textcolor{green}{14.59}} & 0.217 & 22.72 & \textbf{\textcolor{red}{25.32}} & \textbf{\textcolor{red}{295.48}}\\
    & W4A16+c & 0.053 & 16.83 & 13.76 & 151.96 & 0.189 & 13.28 & 10.87 & 15.51  & 0.210 & \textbf{\textcolor{green}{33.28}} & 10.58 & 266.76 \\
    & Mini+c & 0.053 & \textbf{\textcolor{green}{17.98}} & 21.08 & 340.85 & \textbf{\textcolor{green}{0.221}} & 17.66 & 18.29 & 33.04 & 0.121 & 26.12 & 18.06 & \textbf{\textcolor{green}{119.06}} \\
    & W4A16+KV\textsuperscript{Q} & 0.022 & \textbf{\textcolor{red}{8.41}} & \textbf{\textcolor{green}{9.47}} & 229.94 & 0.128 & 10.98 & 9.47 & 49.16 & \textbf{\textcolor{green}{0.218}} & 21.53 & 9.47 & 275.55  \\
    & Mini+KV\textsuperscript{Q} & \textbf{\textcolor{red}{0.013}} & 9.84 & 16.89 & \textbf{\textcolor{red}{493.67}} & 0.153 & 14.77 & 16.87 & 237.86 & 0.115 & \textbf{\textcolor{red}{18.84}} & 16.85 & 229.00 \\
    & Mini+W4A16 & 0.014 & 14.65 & 14.96 & 492.26 & \textbf{\textcolor{red}{0.126}} & \textbf{\textcolor{green}{21.01}} & \textbf{\textcolor{green}{9.45}} & \textbf{\textcolor{red}{239.77}} & \textbf{\textcolor{red}{0.110}} & 26.89 & \textbf{\textcolor{green}{9.09}} & 230.05\\
    \bottomrule
    
\end{tabular} }
 \caption{Evaluation Results for Level 2 Optimization}
 \label{tab:all_lvl2_metrics}
\end{table*}

\textbf{Level 2 optimizations illustrate distinct performance-resource tradeoffs, highlighted by KV-Q+c's balanced accuracy and W4A16+c's superior resource efficiency}. Transitioning from level 1 to level 2 optimizations reveals clear tradeoffs between performance and resource usage. KV-Q+c achieves the highest level 2 average score (18.94\%), a modest 4.87\% drop from the baseline (19.91\%), with limited resource gains (1.11x memory reduction and 1.08x throughput increase), showing notable strength on the 2WikiMQA task. Conversely, W4A16+c, despite a larger score reduction of 11.5\% (score of 18.01), offers substantially better resource efficiency (2.3x memory reduction, 1.6x throughput increase). Although W4A16+c underperforms in SDQA and MDQA tasks, its strong performance in summarization makes it particularly advantageous for processing extensive texts like scientific literature.

\textbf{Combining weight quantization with KV quantization achieves top summarization performance with minimal quality loss but increased token generation}. W4A16+KV\textsuperscript{Q} performed best for summarization tasks. As W4A16 already excelled in summarization tasks, the addition of KV quantization resulted in minimal degradation in text generation quality. This demonstrates that both quantization techniques synergize well, though this combination generates more tokens than other approaches. This combination of weight and KV quantization results in a modest decrease in score; however, this resulted in the highest number of tokens generated.

\begin{table*}
\centering
\resizebox{\textwidth}{!}{%
\begin{NiceTabular}{|m{0.1cm}|m{2.0cm}|>{\centering\arraybackslash}m{1.5cm}|>{\centering\arraybackslash}m{1.3cm}|>{\centering\arraybackslash}m{1.4cm}|>{\centering\arraybackslash}m{2.0cm}|>{\centering\arraybackslash}m{1.5cm}|>{\centering\arraybackslash}m{1.3cm}|>{\centering\arraybackslash}m{1.4cm}|>{\centering\arraybackslash}m{2.0cm}|}
\toprule
\Block{1-*}{}
   & &\multicolumn{4}{c}{\textbf{Qasper}} & \multicolumn{4}{c}{\textbf{HotpotQA}} \\
   \cmidrule[\heavyrulewidth](lr){1-2} \cmidrule[\heavyrulewidth](lr){3-6} \cmidrule[\heavyrulewidth](lr){7-10}
    & \textbf{Method} & F1 QA $\uparrow$ & Prec $\uparrow$ & Recall $\uparrow$ & Hall. Score $\downarrow$ & F1 QA $\uparrow$ & Prec $\uparrow$ & Recall $\uparrow$ & Hall. Score $\downarrow$ \\
   \cmidrule(lr){1-2} \cmidrule(lr){3-6} \cmidrule(lr){7-10}
   \parbox[t]{0mm}{\multirow{7}{*}{\rotatebox[origin=c]{90}{Llama3.1-8B}}}    
    & FP16 & 14.41 & 9.22 & 62.67 & 18.98 & 11.59 & 6.84 & 66.04 & 11.52 \\
    & KV-Q+c & \cellcolor{red!40}{10.92} & \cellcolor{red!40}{6.68} & 58.40 & 17.76 & 11.40 & 6.74 & 62.41 & 9.69 \\
    & W4A16+c & 11.21 & 6.93 & 57.58 & 16.84 & \cellcolor{red!40}{11.06} & 6.56 & 61.00 & 10.42\\
    & Mini+c & 22.09 & \textbf{\cellcolor{blue!15}{24.22}} & \cellcolor{red!40}{35.03} & \textbf{\cellcolor{blue!15}{11.76}} & \textbf{\cellcolor{blue!15}{34.41}} & \textbf{\cellcolor{blue!15}{35.57}} & 47.28 & \textbf{\cellcolor{blue!15}{4.93}}\\
    & W4A16+KV\textsuperscript{Q} & 13.76 & 8.65 & \textbf{\cellcolor{blue!15}{62.22}} & \cellcolor{red!40}{18.78} & 11.09 & \cellcolor{red!40}{6.49} & \textbf{\cellcolor{blue!15}{67.54}} & 12.27 \\
    & Mini+KV\textsuperscript{Q} & \textbf{\cellcolor{blue!15}{24.50}} & 22.62 & 43.90 & 17.57 & 34.16 & 35.08 & 44.35 & 17.82 \\
    & Mini+W4A16 & 21.42 & 19.12 & 40.30 & 17.96  & 24.95 & 24.75 & \cellcolor{red!40}{40.07} & \cellcolor{red!40}{20.56}\\
    \cmidrule(lr){1-2} \cmidrule(lr){3-6} \cmidrule(lr){7-10}

    \parbox[t]{0mm}{\multirow{7}{*}{\rotatebox[origin=c]{90}{Mistral-Nemo-12B}}}
    & FP16 & 12.92 & 8.05 & 54.55 & 22.21 & 8.45 & 5.19 & 53.69 & 16.86 \\
    & KV-Q+c & 12.96 & 8.06 & \textbf{\cellcolor{blue!15}{55.22}} & 18.98  & 13.62 & 9.71 & \textbf{\cellcolor{blue!15}{57.81}} & \textbf{\cellcolor{blue!15}{11.24}} \\
    & W4A16+c & \cellcolor{red!40}{12.66} & 8.08 & \cellcolor{red!40}{47.73} & 19.15 & 10.93 & 6.96 & 54.07 & 11.50 \\
    & Mini+c & \textbf{\cellcolor{blue!15}{18.63}} & \textbf{\cellcolor{blue!15}{14.28}} & 49.04 & \textbf{\cellcolor{blue!15}{18.64}} & \textbf{\cellcolor{blue!15}{28.02}} & \textbf{\cellcolor{blue!15}{27.04}} & 57.05 & 19.73  \\
    & W4A16+KV\textsuperscript{Q} & 12.78 & \cellcolor{red!40}{7.82} & 54.39 & 21.33 & \cellcolor{red!40}{7.24} & \cellcolor{red!40}{4.20} & 49.78 & 16.53\\
    & Mini+KV\textsuperscript{Q} & 18.06 & 13.73 & 48.91 & \cellcolor{red!40}{28.80} & 13.37 & 12.82 & 33.61 & 46.51\\
    & Mini+W4A16 & 16.98 & 12.79 & 48.79 & 28.58 & 12.31 & 10.82 & \cellcolor{red!40}{31.19} & \cellcolor{red!40}{47.15}\\
    \bottomrule 
    & \textbf{Method} & \multicolumn{4}{c}{\textbf{NarrativeQA}} & \multicolumn{4}{c}{\textbf{2WikiMQA}} \\
    \cmidrule(lr){1-2} \cmidrule(lr){3-6} \cmidrule(lr){7-10}
    \parbox[t]{0mm}{\multirow{7}{*}{\rotatebox[origin=c]{90}{Llama3.1-8B}}}
    & FP16 & 9.57 & 6.20 & 44.96 & 22.75 & 29.18 & 25.83 & 42.33 & 92.46\\
    & KV-Q+c & 10.35 & 6.55 & 43.62 & 18.59 & \textbf{\cellcolor{blue!15}{29.69}} & \textbf{\cellcolor{blue!15}{28.15}} & 35.89 & 88.70\\
    & W4A16+c & \textbf{\cellcolor{blue!15}{10.56}} & 6.81 & \textbf{\cellcolor{blue!15}{47.77}} & 18.18  & 25.17 & 23.83 & \cellcolor{red!40}{31.15} & 90.83 \\
    & Mini+c & 8.83 & \textbf{\cellcolor{blue!15}{6.86}} & 23.05 & \textbf{\cellcolor{blue!15}{15.61}}  & 21.42 & 19.57 & 37.69 & \textbf{\cellcolor{blue!15}{8.86}} \\
    & W4A16+KV\textsuperscript{Q} & 8.88 & 5.63 & 44.41 & 22.98 & \cellcolor{red!40}{20.87} & \cellcolor{red!40}{17.34} & \textbf{\cellcolor{blue!15}{41.19}} & \cellcolor{red!40}{93.47}\\
    & Mini+KV\textsuperscript{Q} & 7.12 & 6.27 & 20.15 & \cellcolor{red!40}{23.20} & 24.78 & 22.89 & 37.40 & 21.65 \\
    & Mini+W4A16 & \cellcolor{red!40}{6.23} & \cellcolor{red!40}{5.27} & \cellcolor{red!40}{17.81} & 21.46 & 22.37 & 19.75 & 38.63 & 24.20 \\
      \cmidrule(lr){1-2} \cmidrule(lr){3-6} \cmidrule(lr){7-10}
    \parbox[t]{0mm}{\multirow{7}{*}{\rotatebox[origin=c]{90}{Mistral-Nemo-12B}}}
    & FP16 & 2.51 & 1.47 & 19.93 & 36.52 & 14.44 & 11.89 & 32.95 & 96.84\\
    & KV-Q+c & \textbf{\cellcolor{blue!15}{6.07}} & 3.65 & \textbf{\cellcolor{blue!15}{37.42}} & \textbf{\cellcolor{blue!15}{24.64}} & 21.99 & \textbf{\cellcolor{blue!15}{19.37}} & 36.06 & \cellcolor{red!40}{96.61}\\
    & W4A16+c & 5.29 & 3.50 & 30.86 & 24.78 & 18.89 & 16.53 & \cellcolor{red!40}{33.09} & 92.69 \\
    & Mini+c & 5.30 & \textbf{\cellcolor{blue!15}{3.94}} & 21.96 & 39.72 & \textbf{\cellcolor{blue!15}{22.14}} & 18.85 & \textbf{\cellcolor{blue!15}{48.79}} & \textbf{\cellcolor{blue!15}{18.61}} \\
    & W4A16+KV\textsuperscript{Q} & 2.22 & 1.32 & 23.51 & 38.41 & 12.84 & 10.24 & 33.10 & 96.22\\
    & Mini+KV\textsuperscript{Q} & \cellcolor{red!40}{1.30} & 0.93 & \cellcolor{red!40}{10.42} & 68.00 & 15.29 & 12.95 & 36.19 & 42.86\\
    & Mini+W4A16 & 1.36 & \cellcolor{red!40}{0.86} & 11.45 & \cellcolor{red!40}{68.01} & \cellcolor{red!40}{12.60} & \cellcolor{red!40}{10.00} & 34.75 & 46.40 \\
\bottomrule
\end{NiceTabular} }
    \caption{Accuracy comparison for level 2. Left column is SDA and right column is MDA.}
     \label{tab:all_lvl2_f1}
\end{table*}

\end{document}